\documentclass[pdflatex,sn-mathphys-num, iicol]{sn-jnl}% Math and Physical Sciences Numbered Reference Style 

\usepackage{graphicx}%
\usepackage{multirow}%
\usepackage{amsmath,amssymb,amsfonts}%
\usepackage{amsthm}%
\usepackage{mathrsfs}%
\usepackage[title]{appendix}%
\usepackage{xcolor}%
\usepackage{textcomp}%
\usepackage{manyfoot}%
\usepackage{booktabs}%
\usepackage{algorithm}%
\usepackage{algorithmicx}%
\usepackage{algpseudocode}%
\usepackage{listings}%
\usepackage{adjustbox}
\usepackage{tabularx} 
\usepackage{longtable}
\usepackage{caption}
\usepackage{array}
\usepackage{verbatim}
\usepackage{float}
\usepackage{pifont}
\usepackage[utf8]{inputenc}
\usepackage{xurl}
\usepackage{tcolorbox}
\usepackage{fancyvrb}
\usepackage{ulem}
\usepackage{enumitem} 

\usepackage{rotating}
\usepackage{array}
\usepackage{booktabs}
\usepackage{colortbl}
\definecolor{lightblue}{rgb}{0.8,0.9,1.0}
\definecolor{lightgray}{rgb}{0.95,0.95,0.95}
\definecolor{mygray}{rgb}{0.42, 0.42, 0.42} 

\definecolor{UNIFIblue}{rgb}{0, 0.298, 0.494}  
\definecolor{backgroundUNIFI}{rgb}{0.921, 0.968, 1}

\usepackage{graphicx}%
\usepackage{multirow}%
\usepackage{amsmath,amssymb,amsfonts}%
\usepackage{amsthm}%
\usepackage{mathrsfs}%
\usepackage[title]{appendix}%
\usepackage{xcolor}%
\usepackage{textcomp}%
\usepackage{manyfoot}%
\usepackage{booktabs}%
\usepackage{algorithm}%
\usepackage{algorithmicx}%
\usepackage{algpseudocode}%
\usepackage{listings}%
\usepackage{adjustbox}
\usepackage{tabularx} 
\usepackage{longtable}
\usepackage{caption}
\usepackage{array}
\usepackage{verbatim}
\usepackage{float}
\usepackage{pifont}
\usepackage[utf8]{inputenc}
\usepackage{xurl}
\usepackage{tcolorbox}
\usepackage{fancyvrb}

\definecolor{UNIFIblue}{rgb}{0, 0.298, 0.494}  
\definecolor{backgroundUNIFI}{rgb}{0.921, 0.968, 1}

\usepackage{tcolorbox}
\usepackage{xcolor}
\usepackage{geometry}
\usepackage{cuted}

\definecolor{lightgray}{gray}{0.95}
\definecolor{darkgray}{gray}{0.2}

\tcbuselibrary{skins}
\newtcolorbox{chatbox}[1][]{
    enhanced,
    colback=backgroundUNIFI,
    colframe=UNIFIblue,
    coltitle=white,
    fonttitle=\bfseries,
    title=#1,
    attach boxed title to top left={xshift=0.5cm,yshift=-0.25cm},
    boxed title style={colback=UNIFIblue, colframe=UNIFIblue},
    sharp corners,
    boxsep=5pt,
    left=5pt,
    right=5pt,
    top=5pt,
    bottom=5pt,
}

\begin{document}

\title[BoundingDocs: a Unified Dataset for Document Question Answering with Spatial Annotations]{BoundingDocs: a Unified Dataset for Document Question Answering with Spatial Annotations}


\author*[1]{\fnm{Simone} \sur{Giovannini}} \email{simone.giovannini1@unifi.it}

\author[2]{\fnm{Fabio} \sur{Coppini}} \email{fabio.coppini@letxbe.ai}
\equalcont{These authors contributed equally to this work.}

\author[2]{\fnm{Andrea} \sur{Gemelli}} \email{andrea.gemelli@letxbe.ai}
\equalcont{These authors contributed equally to this work.}

\author[1]{\fnm{Simone} \sur{Marinai}} \email{simone.marinai@unifi.it}
\equalcont{These authors contributed equally to this work.}

\affil[1]{\orgdiv{DINFO}, \orgname{Università degli Studi di Firenze}, \orgaddress{\street{Via di Santa Marta, 3}, \city{Florence}, \postcode{50139}, \country{Italy}}}

\affil[2]{\orgname{LETXBE}, \orgaddress{\street{229 Rue Saint-Honoré}, \city{Paris}, \postcode{75001}, \country{France}}}

\abstract{We present a unified dataset for document Question-Answering (QA), which is obtained combining several public datasets related to Document AI and visually rich document understanding (VRDU). Our main contribution is twofold: on the one hand we reformulate existing Document AI tasks, such as Information Extraction (IE), into a Question-Answering task, making it a suitable resource for training and evaluating Large Language Models; on the other hand, we release the OCR of all the documents and include the exact position of the answer to be found in the document image as a bounding box. Using this dataset, we explore the impact of different prompting techniques (that might include bounding box information) on the performance of open-weight models, identifying the most effective approaches for document comprehension.}


\maketitle

\renewcommand{\thefootnote}{}
\footnotetext{Accepted for publication in the International Journal of Document Analysis and Recognition (IJDAR).}
\addtocounter{footnote}{0}

\section{Introduction}\label{ch:chapter1}

The increasing number of documents produced in various fields, including scientific research, legal proceedings, healthcare, and business, has created an enormous demand for efficient information extraction (IE) methods.

In document processing research, Optical Character Recognition (OCR) has proven essential for transforming scanned documents and images into machine-readable text, facilitating further analysis. Initially, statistical methods\ \cite{mihalcea-tarau-2004-textrank} were used alongside OCR to extract information, followed by machine learning approaches. Subsequently, deep learning techniques\ \cite{DBLP:journals/corr/abs-1810-04805}, especially methods related to Natural Language Processing (NLP), became crucial in advancing document understanding. Today, the focus has shifted towards Large Language Models (LLMs)\ \cite{DBLP:journals/corr/abs-2303-08774}, which, with their exceptional ability to model natural language in complex contexts, have further enhanced document comprehension and the automation of information extraction from extensive volumes of text.

OCR tools and LLMs are now extensively used to perform several tasks in Document AI, including:
\begin{itemize}
    \item \textbf{Document Image Classification}: classifies document images into types such as invoices, scientific papers, and receipts\ \cite{docimageclass};
    \item \textbf{Visual Information Extraction}: extracts entities and relationships from unstructured content, considering text, visual elements, and layout\ \cite{DBLP:conf/aaai/WangLJT0ZWWC21};
    \item \textbf{Visual Question Answering}: answers natural language questions based on a document’s content\ \cite{docvqa}.
\end{itemize}

The two main motivations for building the \texttt{BoundingDocs} dataset\footnote{The dataset is publicly available at \url{https://huggingface.co/datasets/letxbe/BoundingDocs}.}, that is focused  on Information Extraction and Question Answering, are:
\begin{enumerate}
    \item the lack of extensive and diverse QA datasets in the field of Document AI;
    \item the lack of precise spatial coordinates in the existing datasets.
\end{enumerate}

Current datasets do not effectively incorporate positional data, which is essential for reducing hallucinations and improving performance by enabling LLMs to understand document layout more precisely. In contrast, the proposed dataset, \texttt{BoundingDocs}, is specifically designed to capture positional information, which not only defines the dataset's core feature but also serves multiple purposes. First, it allows for verifying whether a model correctly extracts both the value and its location, providing deeper insights into the model’s comprehension of document structure and helping to mitigate hallucinations. Second, it can be leveraged to enhance prompting by incorporating more detailed layout-aware instructions, further improving the model’s ability to interpret and generate structured outputs.

\subsection*{Contribution} \label{sec:rqs}

In this work, we propose a unified approach to build a Question-Answering dataset. Such a dataset can be used for evaluating how good Document AI models are to extract relevant information when answering to natural language questions. In doing so, we aim to address the following research questions:

\begin{itemize}
    \item \textbf{RQ1:} How can existing datasets be unified into a common Question-Answering format?
    \item \textbf{RQ2:} Can rephrased questions generated by LLMs enhance  answer accuracy for document-based questions?
    \item \textbf{RQ3:} Does including layout information in prompts (e.g.\ \cite{DBLP:journals/corr/abs-2306-00526, DBLP:conf/icdar/LamottWUSKO24}) improve the model’s performance on document comprehension tasks?
\end{itemize}

To explore these questions, our study is organized into the following sections. Section\ \ref{ch:chapter2}  reviews the existing literature and benchmarks in the field of Document AI and question answering tasks; Section\ \ref{ch:chapter3}  describes the process of unifying datasets into a common Question-Answering format with enhanced layout annotations; Section\ \ref{ch:chapter4}  evaluates the performance of LLMs using various prompting techniques and presents the results. Conclusions are drawn in Section\ \ref{ch:chapter5}   where we discuss our findings, the key challenges encountered, and  propose directions for future research.

\section{State of the art}\label{ch:chapter2}

We provide an overview of the main models and techniques proposed for Question Answering (QA) and Visual Question-Answering (VQA) \cite{docvqa}. We also discuss the features of the main datasets  in the Document AI that we considered in our research. %, at the best of our knowledge.

\subsection{Related datasets}

As summarized in Table~\ref{tab:dataset_review}, we selected datasets that best match our focus on comprehensive document understanding and advanced VQA, addressing challenges across both single-page and multi-page documents. One detailed review covering more datasets focused on layout-related tasks can be found in \cite{gemelli2024}.

\begin{table*}[ht]
    \centering
    \small
    \resizebox{\textwidth}{!}{%
        \begin{tabular}{@{}l rccccc@{}}
        \toprule
        & \multicolumn{4}{c}{Dataset review} \\ \cmidrule{2-5} & &  \\ 
        {\bf Dataset} & {\bf Size} & {\bf Answers} & {\bf OCR Info} & {\bf OCR Engine}   & {\bf Type} & {\bf Lang} \\\midrule
        VRDU \cite{vrdu} & 2,556 & Yes & 1 & 0 & 1 & 1 \\
        Deepform \cite{deepform} & 60,000 & Yes & 1 & 1 & 1 & 1 \\
        DUDE \cite{dude} & 4,974 & Yes & 1 & 1, 2, 4 & 1 & 1 \\
        FATURA \cite{fatura} & 10,000 & Yes & 2 & 5 & 2 & 1 \\
        SP-DocVQA \cite{docvqa} & 12,767 & No & 1 & 3 & 1 & 1 \\
        MP-DocVQA \cite{TITO2023109834} & 5,929 & No & 1 & 2 & 1 & 1 \\
        FUNSD \cite{funsd} & 199 & Yes & 1 & 0 & 1 & 1 \\
        Kleister Charity \cite{kleister} & 2,788 & No & 3 & 1, 2 & 1 & 1 \\
        Kleister NDA \cite{kleister} & 540 & No & 3 & 1, 2 & 1 & 1 \\
        SROIE \cite{sroie} & 1,000 & No & 1 & 0 & 1 & 1 \\
        XFUND \cite{xfund} & 1,393 & Yes & 1 & 0 & 1 & 2,3,4,5,6,7,8 \\
        SynthTabNet \cite{synthtabnet} & 600,000 & Yes & 1 & 5 & 2 & 1 \\
        CORD \cite{cord} & 1,000 & Yes & 2 & 0 & 1 & 9 \\
        GHEGA \cite{ghega} & 246 & Yes & 1 & 0 & 1 & 10 \\
        \bottomrule
        \end{tabular}%
    }
    \caption{Datasets review details. For clarity, the following codes are used in the table: \textbf{OCR info} - 1: \textit{Full text with bboxes}, 2: \textit{Partial text with bboxes}, 3: \textit{Full text without bboxes}; \textbf{OCR engine} - 0: \textit{Not specified}, 1: \textit{Tesseract}, 2: \textit{Amazon Textract}, 3: \textit{Microsoft OCR}, 4: \textit{Azure Cognitive Service}, 5: \textit{Synthetic document (OCR not needed; text is pre-known)}; \textbf{Type} - 1: \textit{Real}, 2: \textit{Synthetic}; \textbf{Lang} - 1: \textit{English}, 2: \textit{Italian}, 3: \textit{French}, 4: \textit{Spanish}, 5: \textit{Chinese}, 6: \textit{German}, 7: \textit{Portuguese}, 8: \textit{Japanese}, 9: \textit{Indonesian}, 10: \textit{Not specified mix}.}
    \label{tab:dataset_review}
\end{table*}

Among the foundational datasets, \texttt{DocVQA}\ \cite{docvqa, TITO2023109834} stands as one of the earliest benchmarks dedicated to VQA on document images, focusing on understanding both textual and layout aspects of documents. Introduced in 2020, \texttt{DocVQA} comprises multiple tasks designed to push the boundaries of document comprehension. The main tasks include answering questions about individual document pages and analyzing multi-page documents — a crucial capability for real-world applications. The \texttt{Single Page}\ \cite{docvqa} subset includes 50,000 questions over 12,767 documents, while the \texttt{Multi Page}\ \cite{TITO2023109834} subset contains 46,436 questions spanning 5,929 documents (covering 47,952 pages in total). These datasets require models to interpret the visual structure of documents and to derive insights that go beyond simple text extraction.

\texttt{DUDE}\ \cite{dude} builds on this foundational work by extending VQA to multi-domain, multi-purpose documents. The dataset provides 5,000 annotated PDF files with 18,700 question-answer pairs across various domains and time frames, making it a unique resource for tasks that integrate Document Layout Analysis with complex, layout-based question answering. Unlike typical QA datasets, \texttt{DUDE} often requires multi-step reasoning, handling both content and structural queries. For instance, questions may include layout-based prompts such as ``\textit{How many text columns are there?}'' or require arithmetic and comparison skills, presenting a challenging dataset for models trained primarily on text-based QA.

In addition to these datasets, several others serve as standard benchmarks and are worth mentioning briefly. \texttt{VRDU}\ \cite{vrdu} includes two corpora—registration forms from the U.S. Department of Justice and ad-buy forms from the FCC—representing templates of varying complexity. The \texttt{FATURA}\ \cite{fatura} dataset provides 10,000 images across 50 templates with imbalanced distributions for fields commonly found in invoices, such as buyer information and total amount, along with bounding box annotations for structured data extraction. \texttt{Kleister}\ \cite{kleister} datasets offer specialized financial reports and legal documents, with \texttt{Kleister Charity} and \texttt{Kleister NDA} addressing entity extraction for key attributes. \texttt{Deepform}\ \cite{deepform} offers approximately 20,000 labeled receipts for political ad purchases with labeled fields for specific political advertising details.

Finally, \texttt{FUNSD}\ \cite{funsd} and \texttt{XFUND}\ \cite{xfund} are form-centric datasets focused on entity linking and key-value extraction in noisy, often multilingual documents. \texttt{FUNSD} includes 199 annotated forms in English, designed for form understanding, while \texttt{XFUND} broadens this to a multilingual setting with documents in seven languages, capturing the diversity of form structures globally.

Recently, several unified datasets similar to \texttt{BoundingDocs} have been published, aiming to integrate multiple document sources or tasks sharing both similarities and differences with our approach.  
\texttt{Docmatix} \cite{huggingface2024docmatix} and \texttt{K2Q} \cite{DBLP:conf/emnlp/ZmigrodSSMNLV24} are Information Extraction datasets containing 2.1 million and 12,000 documents, respectively. Both generate question-answer pairs automatically using LLMs. Notably, \texttt{K2Q}, which is only available upon request, implements a similar strategy to ours by augmenting questions that were initially created using fixed templates at the dataset level. However, neither dataset includes information about the position of the answer within the text, which distinguishes \texttt{BoundingDocs}.

Additionally, recently published datasets such as \texttt{BigDocs} \cite{DBLP:journals/corr/abs-2412-04626} and \texttt{DocStruct4M} \cite{hu2024mplugdocowl} aim to aggregate multiple tasks beyond Information Extraction. These datasets contain 7 million multimodal documents and 4 million documents, respectively. While significantly larger than \texttt{BoundingDocs}, they fall outside our focus on business documents. Moreover, they encompass a wide range of tasks, including \textit{Screenshot2HTML}, \textit{Table2LaTeX}, \textit{ChartParsing}, and \textit{TableParsing}. Not all examples in these datasets follow a question-answering format, and, most notably, they do not provide bounding box information for the answers.

\subsection{Related methods}

In recent years, the QA task\ \cite{docvqa, TITO2023109834} has been approached in many ways, leveraging different techniques and various model architectures. These methods can be broadly categorized into \textit{NLP-based}, \textit{LLM-based}, and \textit{multimodal architectures}, each addressing different aspects of document understanding and question answering.

\textit{NLP-based approaches} build on general Question-Answering models, primarily focusing on text semantics without explicitly incorporating document layout or visual features. A prime example is \texttt{BertQA}\ \cite{docvqa}, which utilizes a BERT architecture followed by a classification head to predict the start and end indices of an answer span. Modifications such as changes in hyperparameters and the introduction of new pre-training tasks have been explored in multiple works\ \cite{Garncarek_2021, liu2019robertarobustlyoptimizedbert}, resulting in improved outcomes.

\textit{LLM-based methods} leverage large language models to perform document understanding tasks by encoding structural and layout information directly into the input. For instance, \texttt{LMDX}\ \cite{perot2023lmdx} incorporates layout information via bounding box coordinates in the prompt, enhancing retrieval precision and reducing hallucinations. \texttt{DocLLM}\ \cite{wang2023docllm}, which builds on the \texttt{LayoutLM} family, includes a specialized pretraining phase focused on structured layout data to improve document layout understanding. In contrast, \texttt{NuExtract}\ \cite{numind2024nuextract} is designed for extracting structured JSON data from documents, using training data derived from the \texttt{Colossal Clean Crawled Corpus}\ \cite{dodge2021documentinglargewebtextcorpora}.

\textit{Multimodal architectures} combine visual and textual features to enhance document comprehension across layout, content, and structure. Among OCR-free methods, \texttt{mPLUG-DocOWL 1.5}\ \cite{hu2024mplugdocowl} integrates a Vision Transformer (ViT)\ \cite{dosovitskiy2021imageworth16x16words} with an LLM for comprehensive Document AI analysis, aligning layout and textual cues effectively without requiring separate OCR stages. Similarly, \texttt{Donut}\ \cite{kim2022ocrfreedocumentunderstandingtransformer} and \texttt{Dessurt}\ \cite{davis2022endtoenddocumentrecognitionunderstanding} operate without OCR preprocessing, directly integrating image and text data for robust document understanding.

In contrast, OCR-dependent models further refine document comprehension by incorporating OCR-based tokens. \texttt{Hi-VT5}\ \cite{TITO2023109834}, for example, combines OCR tokens with visual features, optimizing its effectiveness for Question-Answering tasks that rely on precise textual information. Additionally, \texttt{LayoutLMv3}\ \cite{huang2022layoutlmv3pretrainingdocumentai} introduces visual patch embeddings in place of traditional CNNs to better align text, layout, and visual cues, resulting in improved performance on tasks requiring fine-grained structural interpretation.

\section{Dataset construction}\label{ch:chapter3}

\begin{figure*}[ht]
    \centering
    \includegraphics[width=\textwidth]{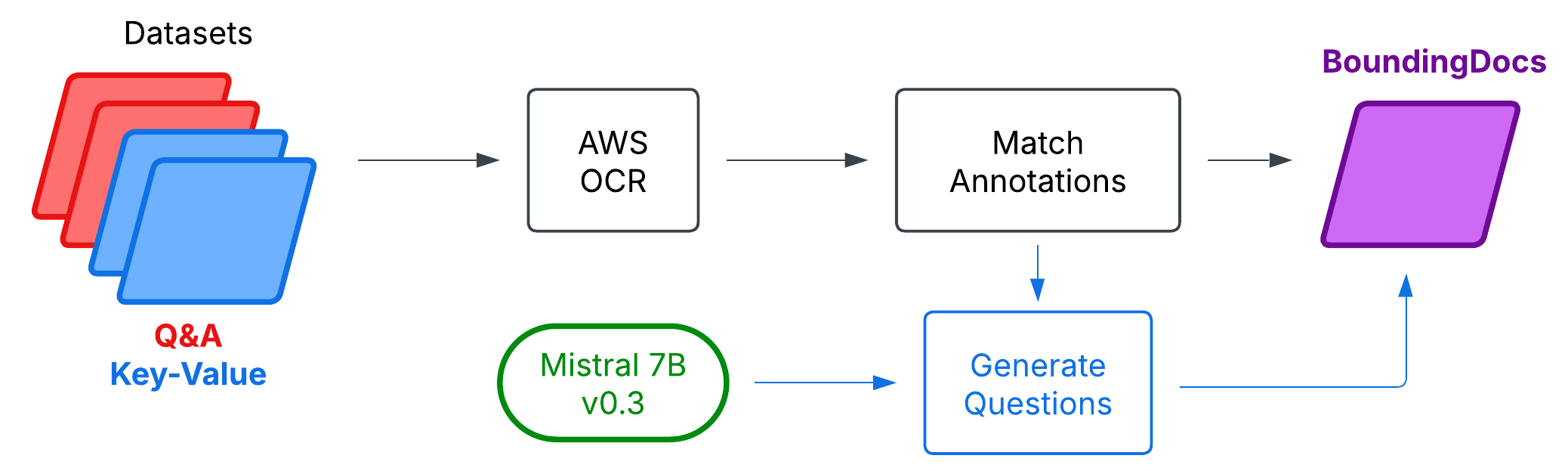}
    \caption{Dataset construction pipeline. The process begins with two main dataset categories: QA datasets (red) and Key-Value extraction datasets (blue). Both categories are processed using AWS OCR (Textract), followed by annotation matching. For the Key-Value extraction datasets, questions are generated and rephrased using \texttt{Mistral 7B v0.3}. All processed components are then unified into \texttt{BoundingDocs} (purple).}
    \label{fig:dataset_construction_pipeline}
\end{figure*}

We base our new dataset, \texttt{BoundingDocs}, on the following datasets selected from Table \ref{tab:dataset_review}: \texttt{SP-DocVQA}, \texttt{MP-DocVQA}, \texttt{DUDE}, \texttt{Deepform}, \texttt{VRDU}, \texttt{FATURA}, \texttt{Kleister Charity}, \texttt{Kleister NDA}, \texttt{FUNSD}, and \texttt{XFUND}. Our selection focuses specifically on business document-based datasets, thereby excluding datasets such as \texttt{InfographicsVQA} \cite{infographicvqa} and \texttt{SlideVQA} \cite{DBLP:conf/aaai/TanakaNNHSS23}, which primarily deal with non-business documents, as well as \texttt{VisualMRC} \cite{DBLP:conf/aaai/TanakaNY21}, which, despite being document-based, features abstractive rather than extractive answers, as detailed in Section \ref{sec:prep}.

Although most of the selected datasets originate from Key-Value extraction tasks, which may involve recurring question types, \texttt{BoundingDocs} ensures diversity by incorporating datasets with heterogeneous characteristics. Specifically, \texttt{BoundingDocs} includes documents with widely varying layouts, structural complexities, and languages, spanning multiple domains such as invoices, contracts, forms, receipts, and documents containing some handwritten-filled forms. This collection provides an essential resource for training and evaluating Document AI models.

In Figure\ \ref{fig:dataset_construction_pipeline} we show the implemented pipeline for dataset construction.

\subsection{Dataset format definition}

For each document, a JSON file contains the annotation (examples  in Figure \ref{fig:qa-pairs-example}). Each word in the answer is linked to its corresponding bounding box. Following established practices in the literature (e.g., \texttt{LayoutLM} \cite{huang2022layoutlmv3pretrainingdocumentai}, \texttt{BERT}\ \cite{DBLP:journals/corr/abs-1810-04805}), the bounding boxes are normalized integers ranging from 0 to 1000 relative to the actual page size. Each bounding box is defined by a list of four values: the width, the height, the $X$ and $Y$ coordinates of the top-left vertex of the rectangle.

\subsection{Producing annotations}

A significant challenge comes from integrating various types of annotations into a unified structure. Datasets like \texttt{Deepform}, \texttt{Kleister}, and \texttt{FATURA} provide annotations that only establish a relationship between a key and its corresponding value in the text, such as annotating \emph{Address = 48 Woodford, SandyFord}. However, these datasets lack essential positional information, such as the text's location, frequency of occurrence, and page number. In contrast, datasets like \texttt{VRDU} and \texttt{MP-DocVQA} provide different types of positional information: the former provides bounding boxes for the values to be extracted, while the latter only specifies which page of the document contains the answer. However, inconsistencies may arise because these datasets utilize different OCR tools, leading to variations in positional measurements and formats. To ensure consistent calculations for bounding box positions, we selected Amazon Textract\ \cite{amazon_textract}, as it is both already deployed in our company's document processing systems (reducing annotation costs and ensuring consistency) and one of the most widely used commercial OCR services.

In the case of \texttt{FUNSD} and \texttt{XFUND}, the datasets contain annotations related only to the text's structure and relationships between elements. Consequently, additional steps are necessary to generate relevant questions from these datasets.

\subsubsection{Dataset preparation} \label{sec:prep}

Upon collecting and downloading the datasets the following preliminary operations have been considered case by case. These additional steps are critical to standardize and prepare the datasets for the generation of annotations. Note that when the test split is not public, as in \texttt{Kleister}, \texttt{DUDE}, and \texttt{DocVQA}, the documents in those test splits and their corresponding annotations are excluded from \texttt{BoundingDocs}.

\textbf{Annotation Conversion}: When the annotations in a dataset have a complex format, they are converted into a standardized, more straightforward format. This is particularly required for the \texttt{VRDU} dataset, where the original annotations require interpretation and conversion.

\textbf{Filtering Pages/Questions}: Some datasets contain redundant or irrelevant content, such as unnecessary pages or questions, which have been removed. For instance, in the \texttt{DocVQA} dataset, pages from the \texttt{Multi Page} set were excluded from the \texttt{Single Page} set to prevent duplication. Additionally, for both \texttt{DUDE} and \texttt{DocVQA} datasets, we filtered out all questions that can be defined as \textit{abstractive}, meaning that the answer is not explicitly present in the document text but instead requires reasoning or synthesis. Since our objective is to provide the exact location of each answer within the document, it is essential to retain only extractive questions, where the answer can be directly found in the text. This ensures that every identified answer has a corresponding position in the document, which would not be possible for abstractive questions.

\textbf{Downloading Original Documents}: In datasets where only annotations are provided without the corresponding documents, the original documents are downloaded from external sources. This step was necessary for the \texttt{Deepform} dataset, where the PDFs were not included alongside the annotations.

\textbf{OCR Processing with Textract}: To ensure consistency across all datasets, Amazon Textract has been applied to all documents, regardless of whether they already contained OCR data. Datasets were processed through Textract not only when OCR data was completely absent, but also when OCR was only provided for the annotated fields. This process has been applied to datasets such as \texttt{VRDU}, \texttt{FATURA}, \texttt{Kleister}, \texttt{SP-DocVQA}, \texttt{Deepform}, \texttt{FUNSD}, and \texttt{XFUND}, where OCR data is either insufficient or not provided.

\textbf{Key-Value Association Creation}: For \texttt{FUNSD} and \texttt{XFUND}, key-value pairs for information extraction were generated from the annotations. This step involves linking elements labeled as questions to their corresponding answers to facilitate coherent information extraction.

\subsubsection{Matching annotations and OCR} \label{3:matching}

To match the answer to each question with the data extracted by Textract\ \cite{amazon_textract}, a script has been developed whose main challenge is to identify the correct word when the same value appears at multiple positions. Our approach matches the annotated value with the extracted text and considers all occurrences as potential matches. While this method ensures broad coverage, it may lead to false positives when the same textual value appears in unrelated contexts. A thorough analysis of this aspect can be found in Section \ref{3:split}.

A considerable time has been devoted to produce high quality annotations. This script, a significant part of our contribution, is used across all datasets with only slight modifications to match the different annotation formats.

For a document and a given key-value pair, where the \textit{key} represents a label (such as ``name,'' ``address,'' or ``date'') describing the type of information, and the \textit{value} contains the actual data associated with that label, the script executes the following steps (no brand new annotations are created):

\begin{figure*}[t]
    \centering
    \begin{minipage}{0.48\textwidth}
        \begin{tcolorbox}[colback=backgroundUNIFI, colframe=UNIFIblue, title=\textbf{Deepform QA pair}]
        \footnotesize
        \begin{Verbatim}[commandchars=\\\{\}]
\textcolor{purple}{"deepform/8385"}: \{
    \textcolor{blue}{"question"}: \textcolor{purple}{"What is the Gross Amount?"},
    \textcolor{blue}{"answers"}: [
        \{
            \textcolor{blue}{"value"}: \textcolor{purple}{"\$576,405.00"},
            \textcolor{blue}{"location"}: [ [\textcolor{purple}{90}, \textcolor{purple}{11}, \textcolor{purple}{364}, \textcolor{purple}{768}] ],
            \textcolor{blue}{"page"}: \textcolor{purple}{1}
        \}
    ],
    \textcolor{blue}{"rephrased_question"}:
        \textcolor{purple}{"What is the value of the Gross Amount?"}
          
\}
        \end{Verbatim}

        \end{tcolorbox}
    \end{minipage}%
    \hspace{0.02\textwidth}
    \begin{minipage}{0.48\textwidth}
        \begin{tcolorbox}[colback=backgroundUNIFI, colframe=UNIFIblue, title=\textbf{Kleister Charity QA pair}]
        \footnotesize
        \begin{Verbatim}[commandchars=\\\{\}]
\textcolor{purple}{"kleister_charity/73938"}: \{
    \textcolor{blue}{"question"}: \textcolor{purple}{"What is the Address Postcode?"},
    \textcolor{blue}{"answers"}: [
        \{
            \textcolor{blue}{"value"}: \textcolor{purple}{"ST4 8AW"},
            \textcolor{blue}{"location"}: [ [\textcolor{purple}{34}, \textcolor{purple}{10}, \textcolor{purple}{692}, \textcolor{purple}{335}] ],
            \textcolor{blue}{"page"}: \textcolor{purple}{1}
        \}
    ],
    \textcolor{blue}{"rephrased_question"}:
        \textcolor{purple}{"What is the postal code of the address?"}
          
\}
        \end{Verbatim}
        \end{tcolorbox}
    \end{minipage}
    \caption{Sample of QA pairs from the dataset. The left QA pair is sourced from \texttt{Deepform}, while the right one is from \texttt{Kleister Charity}. The \textcolor{purple}{purple} values represent the specific details related to each QA pair, and the \textcolor{blue}{blue} keys denote the fixed structure defined for our dataset.}

    \label{fig:qa-pairs-example}
\end{figure*}

\begin{enumerate}
    \item Compare each text line extracted by Textract (\texttt{Line}) with the correct answer using Jaccard similarity. The Jaccard similarity between two sets \( A \) and \( B \) is given by:
    $   J(A, B) = \frac{|A \cap B|}{|A \cup B|}$
    where \( |A \cap B| \) is the number of common elements between the two sets, and \( |A \cup B| \) is the total number of unique elements across both sets.
    \item If similarity exceeds a given threshold, the \texttt{Line} is added to a set of candidates. 
    \item For each candidate line that exceeds the Jaccard similarity threshold, we verify that each word of the ground truth answer is also detected as a \texttt{Word} block by Textract and falls within the \texttt{Line} bounding box. The set of all \texttt{Word} blocks (and their corresponding bounding boxes) that, when concatenated, reconstruct the original answer becomes the localized and annotated answer in \texttt{BoundingDocs}. If the answer is found in multiple locations on the page through this procedure, all occurrences are considered valid answers.
\end{enumerate}

Note that all date annotations in the \texttt{Kleister} datasets follow the \emph{'YYYY-MM-DD'} format. If a date appears in a different format within the document, its original annotation is standardized to this format. Our script is designed to handle this specific case through the use of regular expressions. This was not necessary for the other datasets.

\subsubsection{Questions formulation}

For datasets that are not originally designed for QA but only for key-value extraction, it is necessary, after matching the answers with the OCR output, to also generate the corresponding questions. Therefore, in addition to the three steps described in the previous section, the following two additional steps are performed:

\begin{enumerate}[start=4]
    \item Questions are generated using the template \texttt{What is the [key name]?} (e.g., \texttt{What is the Address?}). Note that \texttt{[key name]} does not refer to the actual name assigned to the key by the datasets (e.g., \texttt{program\_desc}) but rather to a refined, natural-language version defined by us (e.g., \texttt{program description}). We manually refined around 10 keys per dataset. This ensures that questions are always in natural language—though not necessarily correct—but consistently composed of meaningful words. For \texttt{XFUND}, the question template was automatically translated to match document languages. Datasets with pre-defined questions (\texttt{DUDE}, \texttt{MP-DocVQA}, \texttt{SP-DocVQA}) used their own questions.
    \item Moreover, for \texttt{VRDU Ad Buy Form}, additional questions are created to account for key-value pairs linked to specific ad programs, such as:
    \begin{itemize}
        \item \texttt{What is the [program\_start\_date] for [program\_desc]?}
        \item \texttt{What is the [program\_end\_date] for [program\_desc]?}
        \item \texttt{What is the [sub\_amount] for [program\_desc]?}
    \end{itemize}
\end{enumerate}

\subsubsection{Rephrasing questions} \label{3:rephrased_questions}

After the previously described steps, the questions for the new dataset are generated. Inspection of these questions, which followed a simple template-based structure, revealed that they are often grammatically incorrect, overly simplistic, and consistently adhered to the same pattern. This raised concerns that fine-tuning an LLM on these questions could introduce bias, potentially leading to poor performance on questions written by humans, which may not follow the template.

To mitigate this issue, we employed the \texttt{Mistral 7B} model \cite{jiang2023mistral} to correct and rewrite the questions, aiming to fix errors and introduce linguistic diversity. Other Mistral models, such as \texttt{Mistral Large} \cite{mistral2024large} and \texttt{Mixtral 8x7B} \cite{mistral2023mixtral}, were also tested, but they produced overly complex, verbose, and unnatural questions.

The prompt for question rewriting included manually written examples to guide the model, with no information about the correct answer to avoid biasing the generation. For example, the question \texttt{What is the Gross Amount?} was rewritten by the LLM as \texttt{What is the value of the Gross Amount?}.

This procedure was applied to most questions in the dataset, adding a new attribute, \texttt{rephrased\_question}. Questions from \texttt{DUDE}, \texttt{MP-DocVQA}, and \texttt{SP-DocVQA} were excluded as they were already human-written.

To ensure quality and limit hallucinations, we iteratively refined the prompt during the design phase and validated the outputs through manual sampling. Additionally, the LLM was instructed to preserve the semantic meaning a from the template questions, using the original answer as a starting point.

\begin{table*}[t]
\centering
\small
\resizebox{0.95\textwidth}{!}{%
\begin{tabular}{lrrrrrr}
\toprule
\textbf{Dataset} & \textbf{Documents} & \textbf{Pages} & \textbf{Questions} & \textbf{Ques./Page} & \textbf{Ques./Doc} \\
\midrule
Deepform & 24,345 & 100,747 & 55,926 & 0.55 & 2.30 \\
DUDE & 2,583 & 13,832 & 4,512 & 0.33 & 1.75 \\
FATURA & 10,000 & 10,000 & 102,403 & 10.24 & 10.24 \\
FUNSD & 199 & 199 & 1,542 & 7.75 & 7.75 \\
Kleister Charity & 2,169 & 47,550 & 8,897 & 0.19 & 4.10 \\
Kleister NDA & 337 & 2,126 & 696 & 0.33 & 2.07\\
MP-DocVQA & 5,203 & 57,643 & 31,597 & 0.55 & 6.07\\
SP-DocVQA & 266 & 266 & 419 & 1.58 & 1.58\\
VRDU Ad Form & 641 & 1,598 & 22,506 & 14.08 & 35.11\\
VRDU Reg. Form & 1,015 & 2,083 & 3,865 & 1.86 & 3.81\\
XFUND & 1,393 & 1,393 & 16,653 & 11.95 & 11.95\\
\midrule
Total & 48,151 & 237,437 & 249,016 & 1.05 & 5.17\\
\bottomrule
\end{tabular}%
}
\caption{Overall statistics of \texttt{BoundingDocs}, divided by source.}
\label{tab:dataset_statistics}
\end{table*}

In Figure \ref{fig:qa-pairs-example} we show one example of the final format of the dataset questions, including the rephrased 
version of the questions.

\subsection{Statistics \& splits \label{3:split}}

The dataset is split into training, validation, and test sets using an 80-10-10 split based on document count, ensuring that all questions related to a single document remain within the same set. Since the dataset comprises diverse sources with varying structures and question types, we propose our own split to ensure a balanced distribution of document layouts and question types across all sets. Table \ref{tab:dataset_statistics} provides an overview of the dataset's size and source distribution, while detailed statistics can be found in the Supplementary Material.

To achieve this balance, documents from each source dataset are sampled separately. Specifically, documents from \texttt{Deepform} are split in an 80-10-10 ratio, followed by documents from \texttt{FATURA}, \texttt{DUDE}, and all other sources. The union of these individual splits yields the final training, validation, and test sets, ensuring that all sets reflect the dataset’s overall diversity.

\begin{figure*}[t]
    \centering
    \fbox{\includegraphics[width=0.8\textwidth]{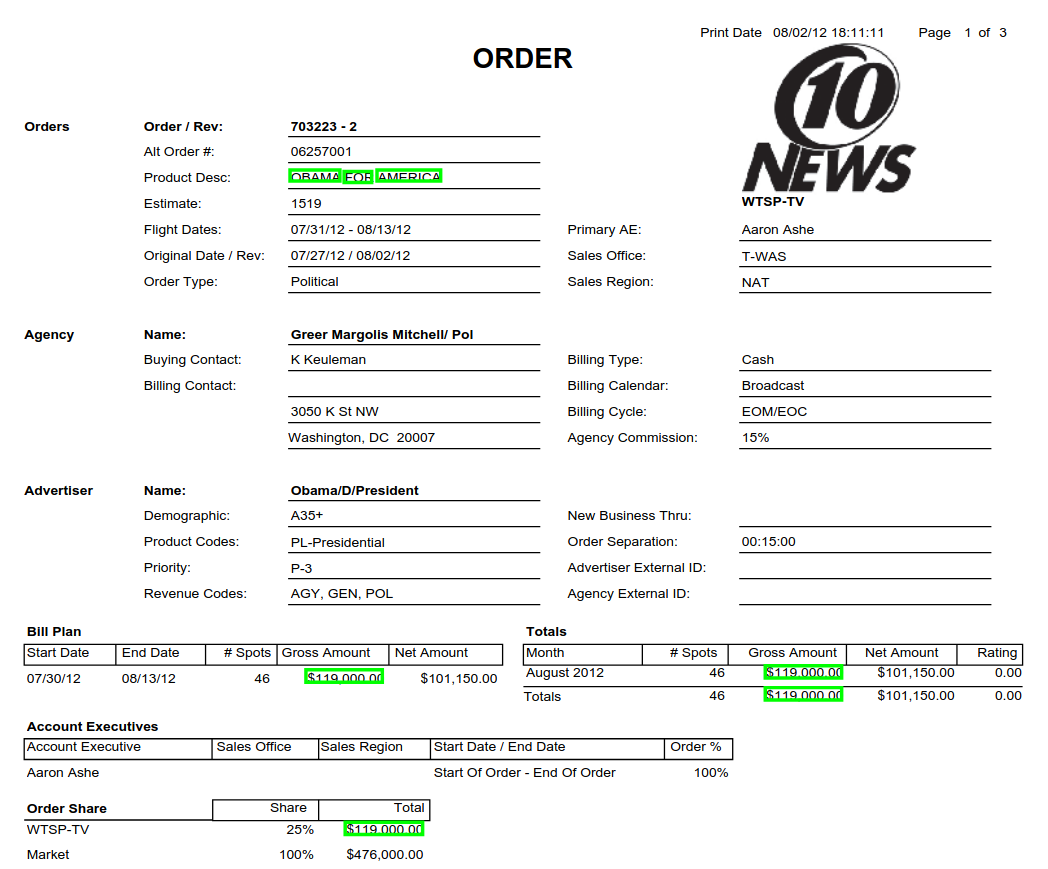}}
    \caption{\texttt{Deepform} page with bbox annotations.}
    \label{fig:deepform_screenshot}
\end{figure*}

\begin{figure*}[t]
    \centering
    \fbox{\includegraphics[width=0.8\textwidth]{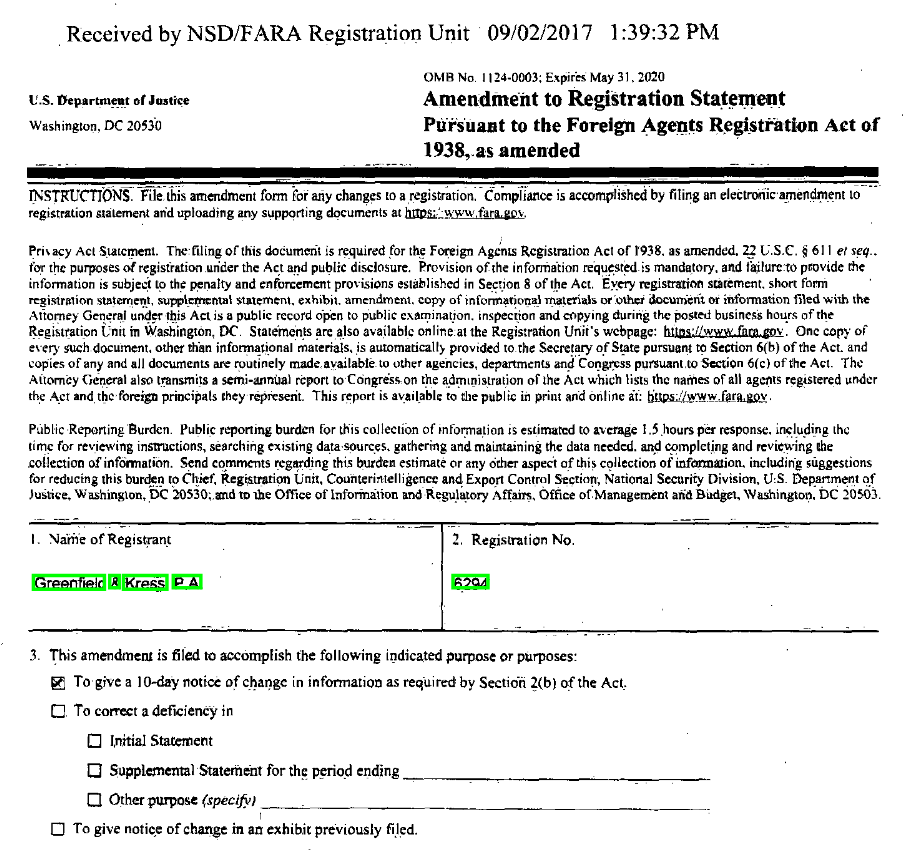}}
    \caption{\texttt{VRDU Registration Form} page with bbox annotations.}
    \label{fig:vrdu_screenshot}
\end{figure*}

Some of the pages annotated using the proposed algorithm and belonging to \texttt{BoundingDocs} are shown in Figures \ref{fig:deepform_screenshot} and \ref{fig:vrdu_screenshot}.
For illustration purpose, colored rectangles are drawn around the fields corresponding to the correct answers to the questions.

It should be noted, as previously mentioned, that for datasets such as \texttt{Kleister}, \texttt{DUDE}, and \texttt{DocVQA}, the test splits are not publicly available. Consequently, documents from these test splits are not included in our dataset.
For datasets like \texttt{FATURA} and \texttt{Deepform}, which do not have predefined splits, this issue does not arise. However, further discussion is needed regarding \texttt{FUNSD}, \texttt{XFUND}, and \texttt{VRDU}, as they do have defined splits.
In the case of \texttt{FUNSD} and \texttt{XFUND}, a portion of the documents from their test splits were included in the random process used to define the splits in \texttt{BoundingDocs}. As a result, some of these documents are part of the \texttt{BoundingDocs} training split. Specifically, out of the $38,515$ documents in the training split, $313$ originate from the \texttt{FUNSD}/\texttt{XFUND} test sets, accounting for approximately $0.8\%$.
Regarding \texttt{VRDU}, its repository provides multiple splits depending on the task of interest, meaning there is no single definitive test split. Therefore, \texttt{VRDU} has been treated similarly to \texttt{FATURA} and \texttt{Deepform}.
This analysis informs users that incorporating \texttt{BoundingDocs} into training data does not prevent models from being evaluated on benchmarks like \texttt{DUDE}, \texttt{DocVQA}, or \texttt{Kleister}.
This leads to the conclusion that \texttt{BoundingDocs} can serve a dual purpose: it can be used as a pretraining tool without significant risk of contamination with established benchmarks, allowing these benchmarks to be employed for evaluating the resulting models. Additionally, \texttt{BoundingDocs} can function as a benchmark itself for assessing models in a question answering setting, even for datasets and document types that were not originally designed for this task.

\subsection{Annotation assessment}

To provide insights into the dataset quality, we analyzed the frequency of multiple matches within a single page, as mentioned in Section \ref{3:matching}. We found that in approximately 20\% of cases, the annotated value appeared more than once on the same page. However, this does not necessarily indicate that 20\% of the dataset contains false positives.
To further investigate this phenomenon, we conducted a manual audit of annotations with multiple matches. We randomly sampled 20 pages containing at least one question with multiple valid answers, selecting one such question per page. We successfully sampled 20 pages for most datasets, with the exception of \texttt{FUNSD} (19 pages), \texttt{SP-DocVQA} (13 pages), and \texttt{FATURA} and \texttt{VRDU Registration Form} (0 pages for both).
For each sampled page-question pair, we manually examined all annotated answer bounding boxes to verify whether all instances referred to the same logical answer or if some were unrelated occurrences of the same textual value. An annotation was considered correct only if all annotated answers corresponded to valid responses to the question, rather than being semantically unrelated occurrences of the same text.
The results, reported in Table \ref{tab:annotation_accuracy}, reveal that approximately 38\% of questions with multiple answers per page contain at least one answer that is not logically related to the question. Extrapolating from this finding, we estimate that approximately 7\% of all annotations (38\% of the 20\% with multiple matches) may exhibit this issue.
Examining the results by data type, we observe distinct patterns in annotation accuracy. Acronyms achieve perfect accuracy (100\%), as they are highly specific and rarely repeated for different purposes within documents. Named entities and currency values show moderate accuracy (68.12\% and 70.00\%, respectively), while dates perform slightly lower at 66.67\%. Numbers exhibit the lowest accuracy rate (46.88\%), as numerical values frequently appear in unrelated sections of documents such as page numbers, reference codes, or unrelated quantities. The ``Other'' category, which includes miscellaneous data types, shows an accuracy of 40.00\%, indicating challenges in handling diverse or less structured information.

\begin{table}[t]
\centering
\small
\begin{tabular*}{\columnwidth}{@{\extracolsep{\fill}}lrrr@{}}
\toprule
\textbf{Data Type} & \textbf{Total} & \textbf{\checkmark} & \textbf{\%} \\
\midrule
Currency & 20 & 14 & 70.00 \\
Date & 18 & 12 & 66.67 \\
Number & 32 & 15 & 46.88 \\
Acronym & 8 & 8 & 100.00 \\
Named entities & 69 & 47 & 68.12 \\
Other & 25 & 10 & 40.00 \\
\midrule
Total & 172 & 106 & 61.63 \\
\bottomrule
\end{tabular*}
\caption{Annotation quality statistics by data type for questions with multiple valid answers per page.}
\label{tab:annotation_accuracy}
\end{table}

\subsection{Dataset examples} \label{3:samples}

\begin{table*}[h]
    \centering
    \footnotesize
    \resizebox{\textwidth}{!}{%
        \begin{tabular}{@{}p{4.5cm} p{4.5cm} p{3cm}@{}}
        \toprule
        {\bf Template Question} & {\bf Rephrased Question} & {\bf Answer} \\\midrule
        What is the Advertiser? & Who is the advertiser? & OBAMA FOR AMERICA \\ \midrule
        What is the Gross Amount? & What is the value of Gross Amount? & \$119,000.00 \\ \midrule
        What is the Registrant Name? & What is the name of the registrant? & Greenfield \& Kress P.A.  \\ \midrule
        What is the Registration Number? & What is the registration number for the company? & 6294 \\
        \bottomrule
        \end{tabular}%
    }
    \caption{QA pairs of the examples in Figure \ref{fig:deepform_screenshot} and Figure \ref{fig:vrdu_screenshot}. The first two refer to the \texttt{Deepform} sample and the last two to the \texttt{VRDU Registration Form} one.}
    \label{tab:triplets}
\end{table*}

\begin{figure*}[]
    \centering
    \includegraphics[width=\textwidth]{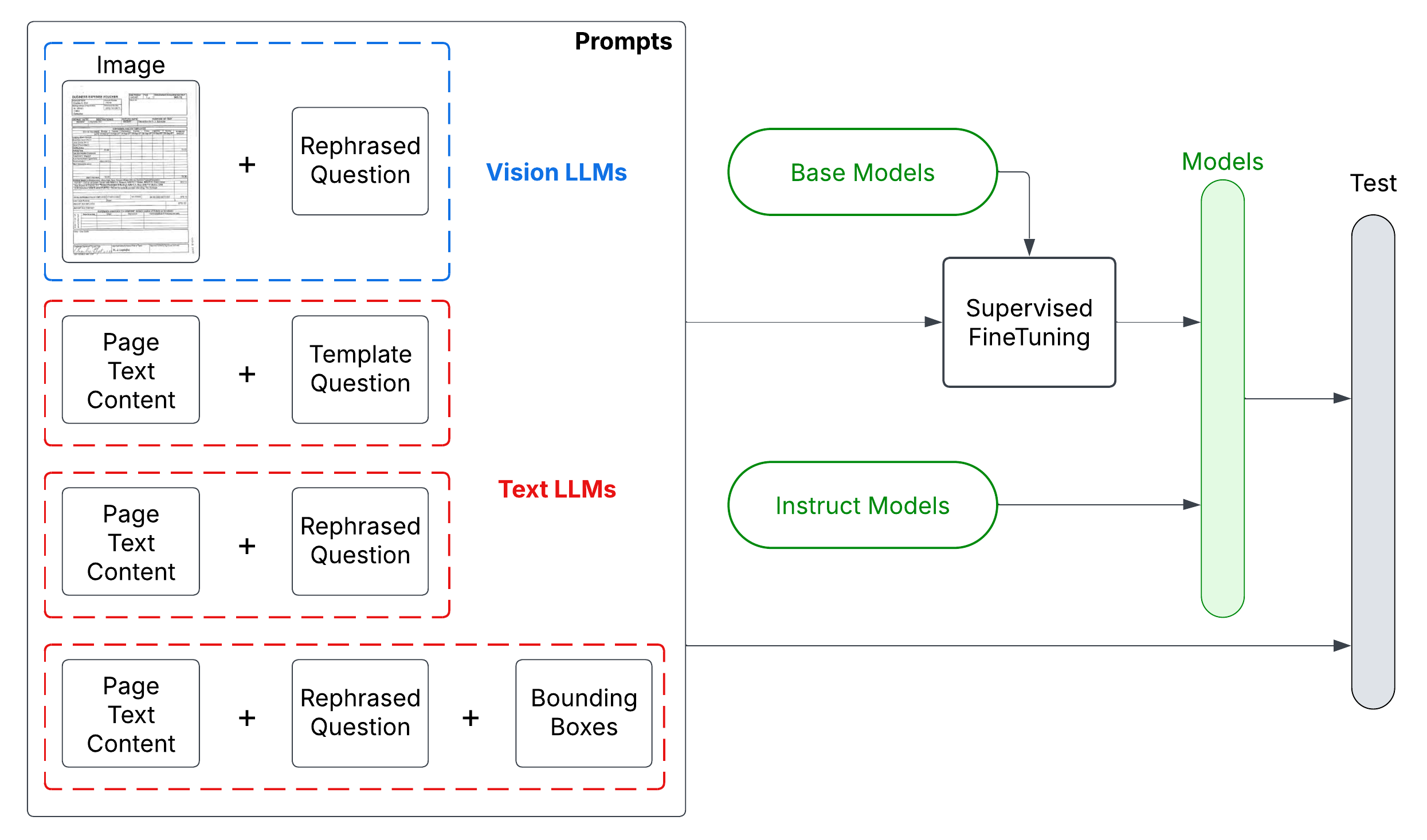}
    \caption{Experimental framework showing the different prompting strategies implemented for vision and text language models. Vision LLMs (blue) are prompted with images and rephrased questions, while text LLMs (red) are prompted with three different configurations of page content and questions. Some models underwent supervised fine-tuning before testing, while others were evaluated directly.}
    \label{fig:testing_pipeline}
\end{figure*}
In Fig. \ref{fig:deepform_screenshot}  (\texttt{Deepform}) and Fig. \ref{fig:vrdu_screenshot} (\texttt{VRDU Registration Form}) it is possible to analyze two pages the corresponding QA pairs are shown in Table \ref{tab:triplets}.
In Fig. \ref{fig:deepform_screenshot}  the extracted fields are the advertiser's name and the gross amount for the various transmissions. In Fig. \ref{fig:vrdu_screenshot} the fields to be extracted are only two: the registrant name and the registration number.

These two examples illustrate how, despite the large number of documents in the collection, the potential amount of information present in the documents is underutilized, as the annotated fields are few compared to the entire body of the documents, indicating that the potential of this large document collection is not being properly exploited.

To provide a concrete response to the issue raised in Section \ref{3:matching}, as shown in Fig. \ref{fig:deepform_screenshot}, the value \texttt{\$119,000.00} has been detected and annotated four times on the page. The two entries on the left part of the page are certainly correct, as they refer to totals: the first is related to the date and the second to the TV transmission. Regarding the values stacked on the right side of the page, the value beneath represents the sum of all those above it, and in this case, the numbers match. This validates the annotation, as the total refers to the same entity. If there were additional entries, only the total would be annotated, as the partial values would necessarily differ.

Additional examples that provide a full overview of the entire variety of the dataset can be found in the Supplementary Material.

\section{Experimental results}\label{ch:chapter4}

Tables \ref{tab:zero-shot}, \ref{tab:ablation}, and \ref{tab:fine-tunings} present our experimental results across the different datasets and model configurations.
The fine-tuning and testing pipeline implemented is summarized and plotted in Figure \ref{fig:testing_pipeline}, which depicts the various types of prompts used depending on the model type: for vision models, we provided the image along with the question only, while for traditional LLMs, we used the OCR-extracted page text, the question, and, for some ablation, the bounding boxes of the text. Notably, all the fine-tuning experiments described were conducted using only 10\% of the training split and evaluated on the proposed test split.

The three tables follow the same format, where the columns correspond to the documents in the \texttt{BoundingDocs} test set derived from the respective dataset. All models and configurations have been evaluated on the proposed test split.
It is crucial to emphasize that these results are not directly comparable to established benchmarks (e.g., \texttt{DocVQA} or \texttt{DUDE}) or to performances reported in other studies using the same datasets. Additionally, comparisons between different model types (e.g., LLMs vs. vision LLMs) should be avoided due to differences in prompting techniques and LoRA fine-tuning parameters (details in the Supplementary Material). The test split is specifically determined by our question-filtering strategy and dataset construction methodology. Therefore, these results primarily serve to illustrate the dataset's difficulty and provide a baseline, rather than to position our approach relative to the state of the art.

\subsection{Evaluation metrics}

We use the standard metric ANLS*\ \cite{peer2024anlsuniversaldocument}, which supports a wider range of tasks including line-item extraction and document-processing tasks. 

For each model-dataset pair in our results, we report two key measurements: the ANLS* value (rescaled between 0 and 100 for easier reading) and the percentage of non-JSON parsable responses relative to the total number of queries. The weighted average provides a comprehensive overview based on the number of examples for each dataset. For ANLS*, higher values indicate better performance, while for non-parsable responses, lower percentages are preferable. In the tables, the best values for each dataset are in bold, whike the second-best values are underlined.

\subsection{Prompt construction}
In this study, each question in the dataset may have answers distributed across multiple pages. Processing multi-page documents poses significant computational challenges, as noted in works such as Multi-PageDocVQA \cite{TITO2023109834}. Additionally, the context size limitations of smaller LLMs make encoding all pages into a single prompt impractical. To address these constraints, we adopt an atomic approach, processing each relevant page separately rather than constructing a single comprehensive prompt.

Our experiments follow the ‘{\it oracle}’ setup described in \cite{TITO2023109834}, where only the page containing the answer is provided as input to the model. This setup isolates the model’s answering capabilities from variations in input sequence length, serving as a theoretical upper bound on performance—assuming the method correctly identifies the relevant page. As a result, these baselines reflect the intrinsic difficulty of the questions and document layouts rather than external factors such as content filtering.
Nevertheless, the dataset includes complete documents, allowing users to implement and evaluate multi-page methods, which must also handle the task of identifying relevant pages.
For questions requiring information from multiple pages, we generate independent prompts for each relevant page, appending the same question to each. For instance, if a five-page document contains relevant information on pages 2 and 4, we create two separate prompts: one incorporating content in page 2 and the other incorporating page 4, both paired with the same question.

Each prompt consists of three components: the document text, the question, and a given answer format (JSON) to facilitate structured data extraction.

\subsection{Baseline models}

% Zero - shot ========================================================

\begin{table*}[htbp]
\centering
\footnotesize
\renewcommand{\arraystretch}{1.5}
\setlength{\tabcolsep}{3pt} 
\rowcolors{2}{white}{lightgray}
\begin{tabular}{l|ccccccccccc|c}
%\hline
\textbf{Model} & \rotatebox{60}{\textbf{Deepform}} & \rotatebox{60}{\textbf{DUDE}} & \rotatebox{60}{\textbf{FATURA}} & \rotatebox{60}{\textbf{FUNSD}} & \rotatebox{60}{\textbf{XFUND}} & \rotatebox{60}{\textbf{SP-VQA}} & \rotatebox{60}{\textbf{Kl. Char.}} & \rotatebox{60}{\textbf{Kl. NDA}} & \rotatebox{60}{\textbf{MP-VQA}} & \rotatebox{60}{\textbf{VRDU Ad}} & \rotatebox{60}{\textbf{VRDU Reg}} & \rotatebox{60}{\textbf{W. Avg.}} \\
\hline
\textbf{Mistral-7B-v0.3} & 42.3 & 9.1 & 6.8 & 14.3 & 6.1 & 22.2 & 21.2 & \underline{32.5} & 12.5 & 23.1 & 28.6 & \cellcolor{lightblue} 22.4 \\
 & 0.22\% & 16.19\% & 1.03\% & 1.23\% & 15.85\% & 10.00\% & 4.84\% & 10.45\% & 7.86\% & 0.22\% & \underline{1.52\%} & \cellcolor{lightblue} 3.32\% \\
\hline
\textbf{Llama-3-8B} & \textbf{83.9} & 60.0 & 35.6 & 70.5 & 38.4 & 73.7 & \underline{72.8} & 25.3 & 62.4 & \underline{71.2} & 37.9 & \cellcolor{lightblue} \underline{62.9} \\
 & 0.47\% & 5.52\% & 0.12\% & 3.68\% & 9.55\% & \underline{2.50\%} & 2.93\% & 6.72\% & 3.54\% & 0.65\% & 6.09\% & \cellcolor{lightblue} 1.77\% \\
\hline
\textbf{Phi-3.5-3.8B} & \underline{66.4} & 45.2 & 24.7 & 55.8 & 51.3 & 50.2 & 63.9 & \textbf{48.1} & 54.4 & 59.2 & 57.9 & \cellcolor{lightblue} 51.6 \\
 & 6.79\% & 64.76\% & 7.40\% & 5.52\% & 2.07\% & 52.50\% & 2.07\% & \textbf{0.00\%} & 52.82\% & 26.80\% & 4.31\% & \cellcolor{lightblue} 20.37\% \\
\hline
\textbf{Claude 3.7 Sonnet} & 62.2 & \textbf{72.4} & \textbf{52.5} & \textbf{78.6} & \underline{54.2} & \underline{85.8} & \textbf{76.8} & 16.6 & \textbf{76.9} & \textbf{90.6} & \textbf{79.6} & \cellcolor{lightblue} \textbf{65.7} \\
 & \textbf{0.00\%} & \textbf{0.00\%} & \textbf{0.00\%} & \underline{1.22\%} & \textbf{0.00\%} & \textbf{0.00\%} & \textbf{0.00\%} & \textbf{0.00\%} & \textbf{0.04\%} & \textbf{0.00\%} & \textbf{0.00\%} & \cellcolor{lightblue} \textbf{0.02\%} \\
\hline
\textbf{Qwen2-VL-2B} & 35.3 & 60.0 & 36.1 & 71.6 & 48.7 & \textbf{88.3} & 48.7 & 28.5 & 68.1 & 52.3 & 62.3 & \cellcolor{lightblue} 45.8 \\
 & 0.57\% & \underline{0.38\%} & \underline{0.05\%} & 1.23\% & 1.41\% & \textbf{0.00\%} & 0.29\% & \underline{1.49\%} & 1.21\% & 1.25\% & \textbf{0.00\%} & \cellcolor{lightblue} \underline{0.54\%} \\
\hline
\textbf{Qwen2-VL-7B} & 44.7 & \underline{69.3} & \textbf{47.5} & \underline{73.2} & \textbf{59.7} & 84.5 & 70.4 & 20.1 & \underline {72.9} & 56.1 & \underline{70.0} & \cellcolor{lightblue} 55.6 \\
 & \underline{0.04\%} & \textbf{0.00\%} & 3.44\% & \textbf{0.00\%} & \underline{0.37\%} & \textbf{0.00\%} & \underline{0.11\%} & \textbf{0.00\%} & \underline{0.14\%} & \underline{0.09\%} & \textbf{0.00\%} & \cellcolor{lightblue} 0.93\% \\
\hline
\end{tabular}
\caption{\textbf{Baseline models}. ANLS* scores and JSON parsing error percentages across datasets for baseline models on our test split. ANLS* scores measure accuracy in answering document questions, while the bottom value in each cell shows JSON parsing errors, indicating output consistency. The "\textbf{W. Avg}" column provides a weighted average across datasets, with bold and underlined values marking the top two scores per dataset.}

\label{tab:zero-shot}
\end{table*}

% ========================================================

We evaluated three popular open-weight models as baselines: \texttt{Mistral 7B Instruct v0.3} \cite{jiang2023mistral}, \texttt{Llama 3 8B Instruct} \cite{touvron2023llama}, and \texttt{Phi 3.5 3.8B Instruct} \cite{abdin2024phi3}. These models were chosen for their established performance and recognition in the NLP community.

In addition to these text-only models, we also tested three multimodal models capable of processing both text and images: \texttt{Claude 3.7 Sonnet} \cite{claudesonnet} and \texttt{Qwen2-VL} \cite{DBLP:journals/corr/abs-2409-12191} in its \texttt{2B} and \texttt{7B} variants. This selection ensures a diverse range of model types and sizes, allowing for a comprehensive evaluation. These tests served to establish an initial benchmark and are reported in Table \ref{tab:zero-shot}.

Through a qualitative analysis based on the results of our preliminary experiments and observations of dataset examples, we identified the most complex layouts and the sources for which the proposed task is particularly challenging. Notably, \texttt{Deepform} and \texttt{VRDU Ad Buy Form} contain numerous similar values that can correspond to different entries in the financial report, complicating information extraction. Despite its seemingly simple layout, \texttt{Kleister NDA} also presents significant challenges. This suggests that the lack of a clear structure — characteristic of contract pages with dense and compact text — impedes the models' ability to retrieve highly specific information. A comprehensive overview of the various sources that compose \texttt{BoundingDocs} can be found in the Supplementary Material.

\subsection{Ablation study: question formulation}

% Ablation on question type ==========================================

\begin{table*}[htbp]
\centering
\footnotesize
\renewcommand{\arraystretch}{1.5}
\setlength{\tabcolsep}{3pt} 
\rowcolors{2}{white}{lightgray}
\begin{tabular}{l|ccccccccccc|c}
%\hline
\textbf{Model} & \rotatebox{60}{\textbf{Deepform}} & \rotatebox{60}{\textbf{DUDE}} & \rotatebox{60}{\textbf{FATURA}} & \rotatebox{60}{\textbf{FUNSD}} & \rotatebox{60}{\textbf{XFUND}} & \rotatebox{60}{\textbf{SP-VQA}} & \rotatebox{60}{\textbf{Kl. Char.}} & \rotatebox{60}{\textbf{Kl. NDA}} & \rotatebox{60}{\textbf{MP-VQA}} & \rotatebox{60}{\textbf{VRDU Ad}} & \rotatebox{60}{\textbf{VRDU Reg}} & \rotatebox{60}{\textbf{W. Avg.}} \\
\hline
\textbf{Templ.-Templ.} & \textbf{97.7} & 70.5 & \textbf{99.9} & \underline{75.7} & 70.1 & 75.3 & 91.9 & \textbf{66.3} & 75.5 & \textbf{96.7} & 96.5 & \cellcolor{lightblue} \underline{91.3} \\
 & \textbf{0.00\%} & \textbf{0.00\%} & \textbf{0.00\%} & \underline{0.61\%} & \textbf{0.61\%} & \textbf{0.00\%} & \textbf{0.00\%} & \textbf{0.00\%} & \textbf{0.06\%} & \textbf{0.00\%} & \textbf{0.00\%} & \cellcolor{lightblue} \textbf{0.04\%} \\
\hline
\textbf{Templ.-Reph.} & 96.8 & 70.9 & 91.5 & 71.1 & 68.8 & 70.2 & 92.5 & 63.7 & 73.6 & 87.1 & 96.0 & \cellcolor{lightblue} 87.8 \\
 & 3.75\% & 1.14\% & \underline{0.23\%} & \textbf{0.00\%} & 1.53\% & \underline{2.50\%} & 0.40\% & 1.49\% & 1.03\% & \underline{0.13\%} & \textbf{0.00\%} & \cellcolor{lightblue} 1.62\% \\
\hline
\textbf{Reph.-Templ.} & \textbf{97.7} & 70.4 & 99.7 & 71.8 & 67.6 & 76.6 & 91.8 & 61.1 & 73.6 & 96.2 & 95.2 & \cellcolor{lightblue} 90.7 \\
 & \textbf{0.00\%} & \textbf{0.00\%} & \textbf{0.00\%} & \textbf{0.00\%} & \underline{0.67\%} & \textbf{0.00\%} & \textbf{0.00\%} & \textbf{0.00\%} & \textbf{0.06\%} & \textbf{0.00\%} & \textbf{0.00\%} & \cellcolor{lightblue} \textbf{0.04\%} \\
\hline
\textbf{Reph.-Reph.} & \underline{97.1} & 71.2 & \underline{99.8} & 72.3 & 68.2 & 76.1 & 92.3 & \underline{64.4} & 73.3 & \underline{96.4} & 95.7 & \cellcolor{lightblue} 90.6 \\
 & \textbf{0.00\%} & \textbf{0.00\%} & \textbf{0.00\%} & \textbf{0.00\%} & 0.92\% & \textbf{0.00\%} & \textbf{0.00\%} & \textbf{0.00\%} & \underline{0.07\%} & \textbf{0.00\%} & \textbf{0.00\%} & \cellcolor{lightblue} \textbf{0.04\%} \\
\hline
\textbf{Reph.-Reph.-bbox} & \textbf{97.7} & \textbf{73.4} & 99.3 & \textbf{78.8} & \textbf{71.2} & \underline{82.1} & \underline{92.8} & 61.6 & \textbf{76.0} & \underline{96.4} & \underline{96.7} & \cellcolor{lightblue} \textbf{91.6} \\
 & 5.01\% & 4.95\% & 5.97\% & 17.79\% & 10.34\% & 5.00\% & 4.34\% & \underline{0.75\%} & 7.16\% & 1.47\% & \underline{0.51\%} & \cellcolor{lightblue} 5.64\% \\
\hline
\textbf{Reph.-Reph.-bbox} & \textbf{97.7} & \underline{72.1} & 99.3 & 74.7 & \underline{70.3} & \textbf{83.0} & \textbf{92.9} & 61.9 & \underline{75.8} & 96.1 & \textbf{96.8} & \cellcolor{lightblue} \underline{91.3} \\
\textbf{w/ regex} & \underline{0.80\%} & \underline{0.38\%} & 4.50\% & 4.29\% & 0.98\% & \textbf{0.00\%} & \underline{0.03\%} & \textbf{0.00\%} & 0.35\% & 0.26\% & \textbf{0.00\%} & \cellcolor{lightblue} \underline{1.53\%} \\
\hline
\end{tabular}
\caption{\textbf{Ablation study}. ANLS* scores and JSON parsing error percentages across datasets for each prompting configuration on our test split. ANLS* scores measure accuracy in answering document questions, while the bottom value in each cell shows JSON parsing errors, indicating output consistency. The "\textbf{W. Avg}" column provides a weighted average across datasets, with bold and underlined values marking the top two scores per dataset.}
\label{tab:ablation}
\end{table*}

% ========================================================

For investigating the impact of question formulation, we selected the \texttt{Mistral 7B v0.3} (base version) for fine-tuning (details in the Supplementary Material). Results are reported in Table \ref{tab:ablation}. We evaluated two types of questions—template-based (simple, consistent format) and rephrased (more varied, user-friendly language). Each model was tested with both question types, resulting in four experimental conditions:

\begin{itemize}
    \item \textbf{Template-Template}: Model trained and tested with template-based questions.
    \item \textbf{Template-Rephrased}: Model trained with template-based questions, tested with rephrased questions.
    \item \textbf{Rephrased-Template}: Model trained with rephrased questions, tested with template-based questions.
    \item \textbf{Rephrased-Rephrased}: Model trained and tested with rephrased questions.
\end{itemize}

\subsection{Incorporating bounding box information}

To assess the impact of spatial information, we incorporated bounding box coordinates into the prompts, denoted as \texttt{Reph.-Reph.-bbox} in Table \ref{tab:ablation}. Each \texttt{Textract}-extracted text element in the prompt was annotated with bounding box coordinates, enabling the model to reference spatial context. In this configuration, the model was specifically fine-tuned to produce more complex JSON outputs that include not only the answer, but also a comprehensive list of all locations where the extracted value appears in the document. While this approach provided richer spatial awareness, the requirement to generate more structured outputs introduced additional complexity that led to increased parsing errors.

To address these parsing challenges, we implemented the \texttt{Reph.-Reph.-bbox w/regex} configuration, which introduced a regex-based post-processing step. When the model's structured JSON output was not parsable due to format inconsistencies or generation errors, the regex extraction mechanism served as a fallback solution to retrieve the target value, effectively maintaining the benefits of spatial information while mitigating the impact of parsing failures.

\subsection{Fine-Tunings}

Table \ref{tab:fine-tunings} presents the fine-tuning results for a subset of the models listed in Table \ref{tab:zero-shot}. In these experiments, models were trained and tested using rephrased questions. For \texttt{Mistral}, the prompt included both the page content and bounding boxes (corresponding to the \texttt{Reph.-Reph.-bbox} row in the Table \ref{tab:ablation}), whereas the two \texttt{Qwen} models received only the page image and the rephrased question as input.  
It is important to noice that these results are not directly comparable across models, as the fine-tuning settings for \texttt{Mistral} and \texttt{Qwen} differ significantly (details on fine-tuning settings can be found in the Supplementary Material.) Once again, these results are not intended for direct comparison, but rather to provide a baseline for the expected performance on this dataset after a fine-tuning of various models.

% Fine Tunings ==========================================

\begin{table*}[htbp]
\centering
\footnotesize
\renewcommand{\arraystretch}{1.5}
\setlength{\tabcolsep}{3pt} 
\rowcolors{2}{white}{lightgray}
\begin{tabular}{l|ccccccccccc|c}
%\hline
\textbf{Model} & \rotatebox{60}{\textbf{Deepform}} & \rotatebox{60}{\textbf{DUDE}} & \rotatebox{60}{\textbf{FATURA}} & \rotatebox{60}{\textbf{FUNSD}} & \rotatebox{60}{\textbf{XFUND}} & \rotatebox{60}{\textbf{SP-VQA}} & \rotatebox{60}{\textbf{Kl. Char.}} & \rotatebox{60}{\textbf{Kl. NDA}} & \rotatebox{60}{\textbf{MP-VQA}} & \rotatebox{60}{\textbf{VRDU Ad}} & \rotatebox{60}{\textbf{VRDU Reg}} & \rotatebox{60}{\textbf{W. Avg.}} \\
\hline
\textbf{Mistral-7B-v0.3} & \textbf{97.7} & \textbf{72.1} & \textbf{99.3} & \underline{74.7} & \textbf{70.3} & \underline{83.0} & \textbf{92.9} & \textbf{61.9} & \textbf{75.8} & \textbf{96.1} & \textbf{96.8} & \cellcolor{lightblue} \textbf{91.3} \\
 & 0.80\% & \underline{0.38\%} & 4.50\% & \underline{4.29\%} & 0.98\% & \textbf{0.00\%} & \textbf{0.03\%} & \textbf{0.00\%} & 0.35\% & \underline{0.26\%} & \textbf{0.00\%} & \cellcolor{lightblue} 1.53\% \\
\hline
\textbf{Qwen2-VL-2B} & 88.0 & 58.9 & 97.2 & 72.9 & 62.4 & 79.7 & 74.1 & \underline{45.6} & 66.1 & 75.9 & 79.9 & \cellcolor{lightblue} 82.1 \\
 & \underline{0.05\%} & \textbf{0.00\%} & \underline{0.30\%} & \underline{0.61\%} & \underline{0.92\%} & \textbf{0.00\%} & \underline{0.43\%} & \underline{0.49\%} & \textbf{0.20\%} & 0.78\% & \textbf{0.00\%} & \cellcolor{lightblue} \underline{0.26\%} \\
\hline
\textbf{Qwen2-VL-7B} & \underline{90.6} & \underline{71.0} & \underline{98.5} & \textbf{75.6} & \underline{68.5} & \textbf{84.2} & \underline{82.8} & 40.9 & \underline{74.9} & \underline{81.2} & \underline{86.0} & \cellcolor{lightblue} \underline{86.6} \\
 & \textbf{0.01\%} & 0.76\% & \textbf{0.00\%} & \underline{0.61\%} & \textbf{0.43\%} & \textbf{0.00\%} & \underline{0.43\%} & \textbf{0.00\%} & \underline{0.28\%} & \textbf{0.00\%} & \underline{0.51\%} & \cellcolor{lightblue} \textbf{0.13\%} \\
\hline
\end{tabular}
\caption{\textbf{Fine-tunings}. ANLS* scores and JSON parsing error percentages across datasets for each fine-tuned model on our test split. ANLS* scores measure accuracy in answering document questions, while the bottom value in each cell shows JSON parsing errors, indicating output consistency. The "\textbf{W. Avg}" column provides a weighted average across datasets, with bold and underlined values marking the top two scores per dataset.}
\label{tab:fine-tunings}
\end{table*}

% ========================================================

\subsection{Answer localization}

In this article, we investigate the unique feature of \texttt{BoundingDocs}, bounding boxes, specifically to evaluate whether their inclusion can enhance document understanding. This can be achieved either by incorporating bounding box information as an additional input in textual prompts for off-the-shelf models, or by leveraging it for fine-tuning. Additionally, \texttt{BoundingDocs} can serve as a benchmark for assessing a model’s ability to accurately predict the location of an answer within a page. Our experiments, reported in \cite{chen2025reliableinterpretabledocumentquestion}, show that \texttt{Claude 4 Sonnet} achieves an average IoU of 0.031 on the \texttt{BoundingDocs} test set, while \texttt{Qwen2.5-VL-7B} reaches 0.048. These results demonstrate that, although off-the-shelf models perform reasonably well in retrieving information in a zero-shot setting, their capacity to indicate the precise location of the answer is effectively nonexistent, which significantly undermines the interpretability of their responses.

% ========================================================

\subsection{Research question answers}

Our experimental findings provide clear answers to our research questions, defined in Section \ref{sec:rqs}:

\begin{itemize}
    \item \textbf{RQ1 - Dataset Unification}: By standardizing data from various sources (e.g., receipts, invoices, forms) into a consistent Question-Answering format, models are exposed to a wide range of document layouts and content types, enhancing their training efficiency. This unification significantly streamlines the fine-tuning process, making it easier to handle diverse document sources. Moreover, fine-tuned models show significant improvements over \texttt{instruct} models, quantitatively confirming that exposure to varied document formats and layouts enhances the model's ability to extract information, as can be seen comparing the overall results in Table \ref{tab:zero-shot} and Table \ref{tab:fine-tunings}.

    \item \textbf{RQ2 - Question Formulation Impact}: The study revealed significant insights into how different question formulation strategies affect document comprehension and information extraction. As reported in Table \ref{tab:ablation}, the \texttt{Template-Template} configuration demonstrated superior performance by leveraging structured, consistent question patterns. However, the \texttt{Rephrased-Rephrased} configuration emerged as a particularly robust solution, maintaining high ANLS* scores (e.g., 99.8 on \texttt{FATURA}, 96.4 on \texttt{VRDU-Ad}) while achieving 0\% parsing errors across most datasets. Crucially, unlike other configurations, its performance remained stable regardless of whether the input followed a template or an augmented version, indicating that the model had developed a greater adaptability to diverse question structures. Notably, the \texttt{Template-Rephrased} setup performed least effectively, highlighting the challenges in transitioning from template-trained models to complex question structures.

    \item \textbf{RQ3 - Layout Information}: The incorporation of spatial information in prompts yielded measurable improvements in model performance, as reported in Table \ref{tab:fine-tunings}. The \texttt{Reph-Reph-bbox} configuration achieved the highest weighted average ANLS* (91.6) across all datasets, demonstrating consistent improvements over configurations without spatial information. Notable gains were observed in complex document understanding tasks, with ANLS* scores increasing to 71.2 on \texttt{XFUND} and 83.0 on \texttt{SP-VQA}. While the initial implementation showed increased parsing errors, the addition of regex-based post-processing (\texttt{Reph-Reph-bbox w/regex}) successfully maintained high performance while reducing error rates to competitive levels (1.53\% weighted average).
\end{itemize}

\section{Conclusions}\label{ch:chapter5}
The paper addresses the growing need for evaluating LLMs in Document AI tasks by proposing a unified dataset designed for document QA taking into account  the position of answers' text in the document. Baseline experiments using off-the-shelf LLMs demonstrate the challenges of applying generic models to specialized Document AI tasks; the performance of instruct models reveal clear limitations in generating correct and well-structured answers.

The paper reveals that while off-the-shelf LLMs struggle with document-specific tasks, targeted fine-tuning can significantly improve their capabilities. Rephrasing questions using LLMs improves the models' understanding and response accuracy across different question formulations, suggesting that LLMs benefit from exposure to diverse linguistic variations during training. Incorporating layout and positional information into the prompt led to improved accuracy across most datasets, but at the cost of a higher percentage of non-parsable responses, reflecting the increased complexity of generating JSON outputs that include bounding box information.

In future work, we aim to explore a wider range of prompting techniques tailored to different model families. Among the many types of information available in \texttt{BoundingDocs}, a key challenge is determining the optimal combination for effective information extraction. Open questions include whether images can fully replace textual content in prompts, whether bounding boxes remain essential even when using images, and how different modalities interact to enhance model performance.
The constructed dataset and the experimental results provide a solid foundation for future research in Document QA. Fine-tuning models with enriched prompts has shown promising improvements.

\section*{Declarations}

\noindent\textbf{Competing interests}\ The authors declare no competing interests.

\bibliography{bibliography}

%=======================================
%Supplementary Material

\appendix

\section{Dataset statistics}

We now provide a quantitative illustration using tables and graphs to show the nature of the dataset in all its aspects.

\begin{figure*}[]
    \centering
    \includegraphics[height=0.45\textheight]{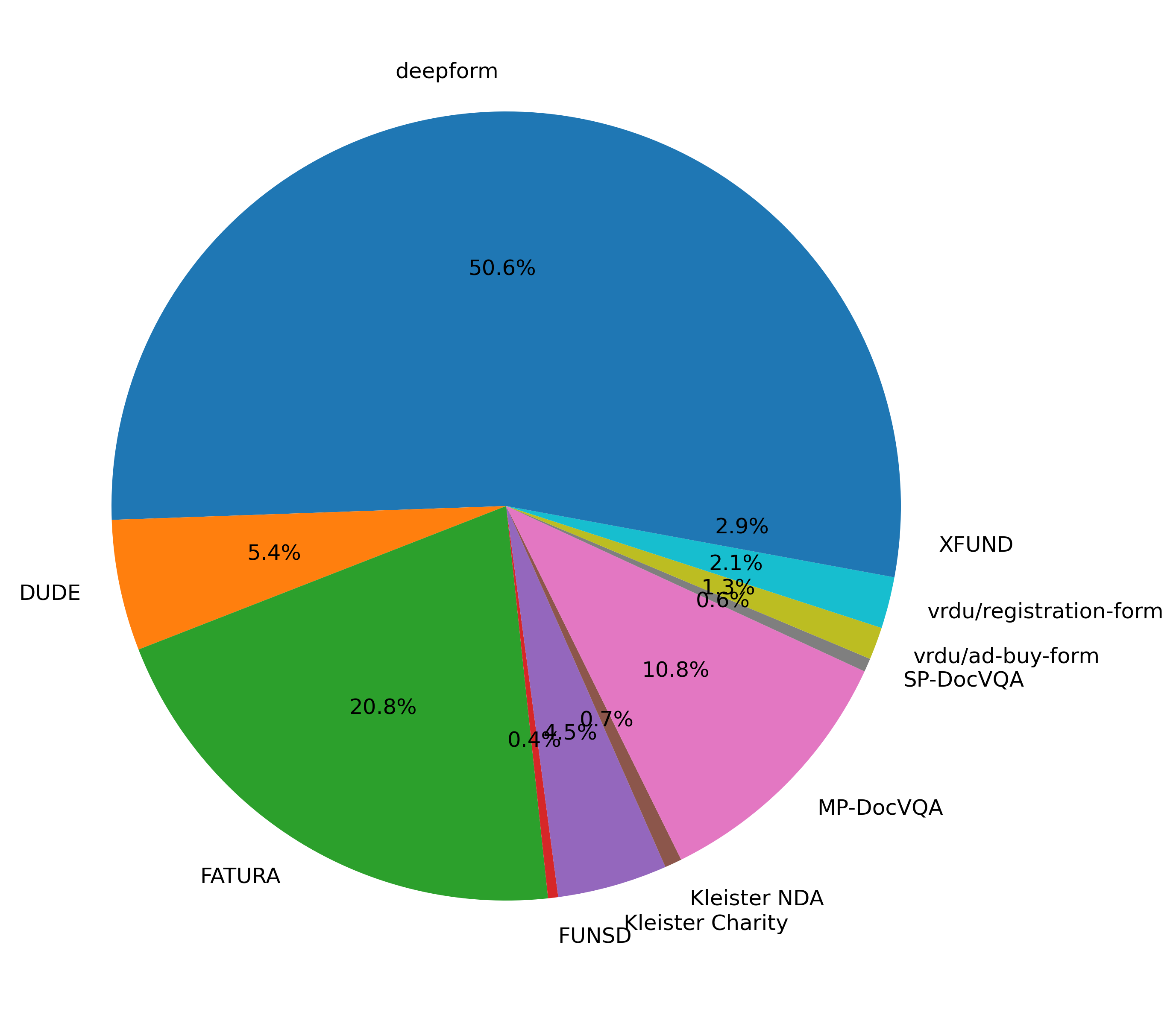}
    \caption{Documents distribution across datasets}
    \label{fig:2_documents_distribution}
\end{figure*}

Figure \ref{fig:2_documents_distribution} provide an overview of the dataset construction, showing how many documents and related questions from the various source datasets contribute to the overall dataset.

There are already several aspects to consider: first of all, it can be seen that \texttt{Deepform} is the dataset that contributes the most documents, but it has an average of about 2 questions per document, whereas the dataset that contributes the most questions is \texttt{FATURA}, with an average of more than 10 questions per document. Note that \texttt{VRDU Ad Buy Form} is the dataset that contains the most annotated fields, and both this aspect and the construction of additional questions for this particular dataset lead to a very high number of questions compared to the relatively low number of documents (an average of more than 34 questions per document).
Also, note that there are very few documents related to \texttt{SP-DocVQA}: this is because, as already mentioned, most of the documents in this dataset were already present in \texttt{MP-DocVQA}, and there was no point in including them twice.

\iffalse
\begin{figure*}[]
    \centering
    \includegraphics[width=\textwidth]{supplementary_img/1_documents_questions_per_dataset.png}
    \caption{Documents and questions per dataset}
    \label{fig:1_documents_questions_per_dataset}
\end{figure*}
\fi

In Table \ref{tab:7_language_distribution} you can observe the distribution of the languages in which the questions in the dataset are posed. The only questions, along with their respective documents, that are not in English are those formulated on \texttt{XFUND}, and thus they represent a clear minority compared to the total count.

\begin{table}[ht]
\centering
\small
\resizebox{\columnwidth}{!}{%
\begin{tabular}{lrr}
\toprule
Language & Questions & Percentage (\%) \\
\midrule
English & 232,362 & 93.31 \\
Italian & 3,857 & 1.55 \\
Spanish & 2,753 & 1.11 \\
French & 2,176 & 0.87 \\
German & 2,564 & 1.03 \\
Portuguese & 3,743 & 1.50 \\
Chinese & 1,116 & 0.45 \\
Japanese & 445 & 0.18 \\
\midrule
Total & 249,016 & 100.00 \\
\bottomrule
\end{tabular}%
}
\caption{Language distribution}
\label{tab:7_language_distribution}
\end{table}

\iffalse
\begin{figure*}[]
    \centering
    \includegraphics[height=0.45\textheight]{supplementary_img/7_language_distribution.png}
    \caption{Language distribution of questions}
    \label{fig:7_language_distribution}
\end{figure*}
\fi

After providing a general overview of the dataset's composition, it is also interesting to conduct an analysis of the types of questions that were generated and which field were extracted by running the matching algorithm on the various source datasets. Obviously, this analysis can only be conducted on the original datasets that pertain to key value extraction, as the questions are constructed according to the previously described template. For datasets such as \texttt{DUDE} and \texttt{DocVQA}, it is not possible to perform this type of tracking.

For \texttt{Deepform}, there are only five fields for which questions have been constructed, as can be observed in Table \ref{tab:9_deepform_question_distribution}. The main fields present are the total cost incurred for the advertisement and the name of the advertiser. The fields Flight From and Flight To are date values that represent the start and end days of the spot's transmission.

\begin{table}[]
\centering
\resizebox{\columnwidth}{!}{
\begin{tabular}{lrr}
\toprule
Question Type & Count & (\%) \\
\midrule
Gross Amount & 15848 & 28.34 \\
Contract Number & 7950 & 14.22 \\
Flight From & 7919 & 14.16 \\
Flight To & 7921 & 14.16 \\
Advertiser & 16288 & 29.12 \\
\midrule
Total & 55926 & 100.00 \\
\bottomrule
\end{tabular}
}
\caption{Question distribution for \texttt{Deepform} dataset}
\label{tab:9_deepform_question_distribution}
\end{table}

\iffalse
\begin{figure*}[]
    \centering
    \includegraphics[height=0.4\textheight]{supplementary_img/9_deepform_question_distribution.png}
    \caption{\texttt{Deepform} question distribution}
    \label{fig:9_deepform_question_distribution}
\end{figure*}
\fi

Regarding \texttt{FATURA}, the range of extracted fields is much broader compared to the previous \texttt{Deepform}, as visible in Table \ref{tab:9_FATURA_question_distribution}. With 50 different layouts within \texttt{FATURA}, not all documents contain the same fields, which is the reason for the significant differences in frequencies among some fields. It is notable that the field with the most questions is Date of purchase (9800), while the least frequent is Total amount to be paid (685).

\begin{table}[]
\centering
\small
\resizebox{\columnwidth}{!}{
\begin{tabular}{lrr}
\toprule
Question Type & Count & (\%) \\
\midrule
Buyer information & 5653 & 5.52 \\
Date of purchase & 9800 & 9.57 \\
Invoice ID & 8796 & 8.59 \\
Remarks and footers & 5157 & 5.04 \\
Seller Address & 8131 & 7.94 \\
Title & 7346 & 7.17 \\
Total amount after tax & 7992 & 7.80 \\
Total words & 3932 & 3.84 \\
GSTIN & 4708 & 4.60 \\
To whom the invoice is sent & 1356 & 1.32 \\
Payment terms and conditions & 2306 & 2.25 \\
Discount & 2383 & 2.33 \\
Due date & 5797 & 5.66 \\
Seller email & 4396 & 4.29 \\
Total amount before tax & 6753 & 6.59 \\
Tax & 3799 & 3.71 \\
Purchase order number & 1400 & 1.36 \\
Total amount to be paid & 685 & 0.67 \\
To whom the bill is sent & 1285 & 1.25 \\
Seller name & 6728 & 6.57 \\
Bank information & 2600 & 2.55 \\
Website of the seller & 1400 & 1.38 \\
\midrule
Total & 102403 & 100.00 \\
\bottomrule
\end{tabular}
}
\caption{Question distribution for \texttt{FATURA} dataset}
\label{tab:9_FATURA_question_distribution}
\end{table}

\iffalse
\begin{figure*}[]
    \centering
    \includegraphics[width=\textwidth]{supplementary_img/9_FATURA_question_distribution.png}
    \caption{\texttt{FATURA} question distribution}
    \label{fig:9_FATURA_question_distribution}
\end{figure*}
\fi

\begin{table}[]
\centering
\small
\resizebox{\columnwidth}{!}{
\begin{tabular}{lrr}
\toprule
Question Type & Count & (\%) \\
\midrule
Charity Name & 1617 & 18.17 \\ 
Charity Number & 2089 & 23.48 \\
Spending Annually in £ & 66 & 0.74\\
Address Post Town & 1948 & 21.90\\
Address Postcode & 1621 & 18.22\\
Address Street Line & 1495 & 16.80\\
Income Annually in £ & 61 & 0.69\\
\midrule
Total & 8897 & 100.00 \\
\bottomrule
\end{tabular}
}
\caption{Question distribution for \texttt{Kleister Charity} dataset}
\label{tab:9_Kleister Charity_question_distribution}
\end{table}

Regarding the \texttt{Kleister Charity} dataset, the range of extracted fields is relatively narrow compared to the \texttt{FATURA} dataset. As shown in Table \ref{tab:9_Kleister Charity_question_distribution}, the dataset primarily focuses on extracting information such as the Charity Name, Charity Number, Address Post Town, Address Postcode, and Address Street Line. The fields with the fewest questions are Spending Annually in British Pounds and Income Annually in British Pounds, indicating that financial details are less frequently extracted from this dataset.

\iffalse
\begin{figure*}[]
    \centering
    \includegraphics[width=\textwidth]{supplementary_img/9_Kleister Charity_question_distribution.png}
    \caption{\texttt{Kleister Charity} question distribution}
    \label{fig:9_Kleister Charity_question_distribution}
\end{figure*}
\fi

\iffalse
\begin{figure*}[]
    \centering
    \includegraphics[width=\textwidth]{supplementary_img/charity_histogram.png}
    \caption{\texttt{Kleister Charity} question distribution}
    \label{fig:9_Kleister Charity_question_distribution}
\end{figure*}
\fi

The \texttt{Kleister NDA dataset}, as detailed in Table \ref{tab:9_Kleister NDA_question_distribution}, contains questions across a very limited set of fields: Jurisdiction, Party, Term, and Effective Date. The field with the most questions is Jurisdiction, followed by Party, while the Effective Date field has only a single question.

\begin{table}[]
\centering
\small
\resizebox{.85\columnwidth}{!}{
\small
\begin{tabular}{lrr}
\toprule
Question Type & Count & (\%) \\
\midrule
Jurisdiction & 319 & 45.83\\
Party & 314 & 45.11\\
Term & 62 & 8.91\\
Effective Date & 1 & 0.15\\
\midrule
Total & 696 & 100.00 \\
\bottomrule
\end{tabular}
}
\caption{Question distribution for \texttt{Kleister NDA} dataset}
\label{tab:9_Kleister NDA_question_distribution}
\end{table}

\iffalse
\begin{figure*}[]
    \centering
    \includegraphics[height=0.4\textheight]{supplementary_img/9_Kleister NDA_question_distribution.png}
    \caption{\texttt{Kleister NDA} question distribution}
    \label{fig:9_Kleister NDA_question_distribution}
\end{figure*}
\fi

\iffalse
\begin{figure*}[]
    \centering
    \includegraphics[height=0.3\textheight]{supplementary_img/kl nda.png}
    \caption{\texttt{Kleister NDA} question distribution}
\end{figure*}
\fi

The \texttt{VRDU Ad Buy Form} dataset, as shown in Table \ref{tab:9_vrdu_ad-buy-form_question_distribution}, contains a broader range of fields compared to the previous datasets. The fields with the most questions are Program Start Date, Channel, and Program End Date. In contrast, the field with the fewest questions is Program Description, with only 6 questions.

\begin{table}[]
\centering
\small
\resizebox{\columnwidth}{!}{
\begin{tabular}{lrr}
\toprule
Question Type & Count & (\%) \\
\midrule
Gross Amount & 614 & 2.72\\
Contract Number & 615 & 2.73\\
Flight From & 439 & 1.95\\
Flight To & 440 & 1.96\\
Advertiser & 584 & 2.59\\
Property & 572 & 2.54\\
Agency & 263 & 1.17\\
Product & 561 & 2.49\\
Sub Amount & 4197 & 18.65\\
Program Start Date & 4714 & 20.95\\
TV Address & 463 & 2.06\\
Channel & 4556 & 20.24\\
Program End Date & 4482 & 19.92\\
Program Description & 6 & 0.03\\
\midrule
Total & 22506 & 100.00 \\
\bottomrule
\end{tabular}
}
\caption{Question Distribution for \texttt{VRDU Ad Buy Form} dataset}
\label{tab:9_vrdu_ad-buy-form_question_distribution}
\end{table}

\iffalse
\begin{figure*}[]
    \centering
    \includegraphics[width=\textwidth]{supplementary_img/9_vrdu_ad-buy-form_question_distribution.png}
    \caption{\texttt{VRDU Ad Buy Form} question distribution}
    \label{fig:9_vrdu_ad-buy-form_question_distribution}
\end{figure*}
\fi

The \texttt{VRDU Registration Form} dataset, as detailed in Table \ref{tab:9_vrdu_registration-form_question_distribution}, contains questions across 6 different fields. The fields with the most questions are Registration Number and Registrant Name, while the field with the fewest questions is Signer Title.

\begin{table}[]
\centering
\small
\resizebox{\columnwidth}{!}{
\begin{tabular}{lrr}
\toprule
Question Type & Count & (\%)\\
\midrule
Registrant Name & 959 & 24.81\\
Registration Number & 983 & 25.43\\
File Date & 783 & 20.25\\
Signer Name & 654 & 16.93\\
Foreign Principle Name & 264 & 6.83\\
Signer Title & 222 & 5.75\\
\midrule
Total & 3865 & 100.00 \\
\bottomrule
\end{tabular}
}
\caption{Question distribution for \texttt{VRDU Registration Form} dataset}
\label{tab:9_vrdu_registration-form_question_distribution}
\end{table}

\iffalse
\begin{figure*}[]
    \centering
    \includegraphics[width=\textwidth]{supplementary_img/9_vrdu_registration-form_question_distribution.png}
    \caption{\texttt{VRDU Registration Form} question distribution}
    \label{fig:9_vrdu_registration-form_question_distribution}
\end{figure*}
\fi

The latest statistics worth noting are those related to the split made for training, validation, and testing. As previously described, the documents were divided according to an 80-10-10 percentage, assuming that the distribution of questions would be similar and that we would therefore obtain the same percentage division for the latter as well. As can be seen from the Table \ref{tab:6_train_val_test_distribution}, our intuition was confirmed, achieving the desired partitioning for the questions as well.

% Table 2: Train/Val/Test Split
\begin{table}[]
\centering
\begin{tabular}{lrrrrr}
\toprule
Split & Docs & Questions & \% Docs & \% Questions \\
\midrule
Train & 38516 & 198601 & 80.0\% & 79.8\% \\
Val & 4804 & 24956 & 10.0\% & 10.0\% \\
Test & 4832 & 25463 & 10.0\% & 10.2\% \\
\bottomrule
\end{tabular}
\caption{Train/Val/Test split}
\label{tab:6_train_val_test_distribution}
\end{table}

\section{Questions rephrasing}

As outlined in the main text, a key method for generating \texttt{BoundingDocs} is question rewriting using LLMs, specifically \texttt{Mistral 7B}. The model was employed through a few-shot approach, as illustrated in Figure \ref{fig:prompt}, where specific examples of the desired question reformulations were provided. The prompt did not include the answer to the question, as we observed that incorporating the answer in the few-shot examples influenced the question formulation. This led to the generation of questions biased toward facilitating information retrieval, thereby compromising the neutrality of the rewriting process.

\begin{figure*}[]
\begin{chatbox}[Questions rephrasing]
\vspace{0.2cm}
\textbf{User:} \\ \\
I have a set of questions written in natural language using a simple template. These questions may contain grammatical errors, unnatural phrasing, or lack richness in expression. Your task is to refine them while keeping their original intent.
\\ \\
--- TASK ---

- Correct any grammatical mistakes

- Improve fluency and clarity

- Enrich phrasing to make the question more natural and well-formed
\\ \\
--- EXAMPLES ---

- Original: [\textit{What is Charity Name?}]
Rephrased: [\textit{What is the name of the Charity?}]

- Original: [\textit{What is the Product?}]
Rephrased: [\textit{What is the name of the product?}]
\\ \\
--- QUESTION TO REPHRASE ---

- Original: [\textit{What is Jurisdiction?}]

\vspace{0.8cm}

\textbf{LLM:} \\ \\ 
\textit{In which state is the company registered?}
\end{chatbox}
\caption{Prompt example and LLM's answer for questions rephrasing.}
\label{fig:prompt}
\end{figure*}

\section{Finetuning details} \label{appendix:finetuning}

For our work, we fine-tuned three different models: \texttt{Mistral-7B-v0.3}, \texttt{Qwen2-VL-2B}, and \texttt{Qwen2-VL-7B}. All three models were fine-tuned on the 10\% of the train split using \texttt{QLoRA}, but with two different configurations: one for \texttt{Mistral}, which processes only text input, and another for the \texttt{Qwen2-VL} models, which also take document images as input. The fine-tuning process was performed in Python using the \texttt{HuggingFace} SFT trainer. Quantization was achieved using the \texttt{BitsAndBytes} library, and \texttt{LoRA} was applied through the \texttt{peft}, both of which are also Python libraries.
\begin{table}[]
\centering
\small
\begin{tabular}{lccc}
\toprule
\textbf{Parameter} & \textbf{Mistral-v0.3} & \textbf{Qwen2-VL} \\
\midrule
\texttt{load\_in\_4bit} & True & True \\
\texttt{use\_double\_quant} & True & True \\
\texttt{quant\_type} & \texttt{"nf4"} & \texttt{"nf4"} \\
\texttt{compute\_dtype} & \texttt{bfloat16} & \texttt{bfloat16} \\
\texttt{lora\_alpha} & 128 & 16 \\
\texttt{lora\_dropout} & 0.05 & 0.05 \\
\texttt{r} & 256 & 8 \\
\texttt{bias} & \texttt{"none"} & \texttt{"none"} \\
\texttt{target\_modules} & \texttt{"all-linear"} & \texttt{"q\_proj"} \\
\texttt{task\_type} & \texttt{"CAUSAL\_LM"} & \texttt{"CAUSAL\_LM"} \\
\bottomrule
\end{tabular}
\caption{Fine-tuning configurations for the \texttt{Mistral} and \texttt{Qwen2-VL} models.}
\label{tab:fine_tuning_configurations}
\end{table}

\section{Dataset examples} \label{appendix:samples}

Similarly to what was done in the paper, a comprehensive qualitative overview of the entire variety of the dataset will be provided. An example will be shown for each source dataset, along with (almost) all the corresponding QA pairs formulated for that page, as shown in Table \ref{tab:app_samples}.

\onecolumn
\begin{center}
    \small
    \begin{longtable}{|p{4cm}|p{4cm}|p{3cm}|p{1cm}|}
        \hline
        \textbf{Template Question} & \textbf{Rephrased Question} & \textbf{Answer} & \textbf{Fig.} \\
        \hline
        \endfirsthead
        \hline
        \textbf{Template Question} & \textbf{Rephrased Question} & \textbf{Answer} & \textbf{Fig.} \\
        \hline
        \endhead
        \hline
        \multicolumn{4}{r}{\textit{Continued on next page}} \\
        \hline
        \endfoot
        \endlastfoot
        What is Advertiser? & Who is the advertiser? & Jordan, Jonathan & \ref{fig:app-deepform-sample} \\
        \hline
        What is Gross Amount? & What is the value of the Gross Amount? & \$10,500.00 & \ref{fig:app-deepform-sample} \\
        \hline
        \addlinespace[5pt] 
        \hline
        What is Address Post Town? & What is the post town of the address? & Stoke-on-Trent & \ref{fig:app-charity-sample} \\
        \hline
        What is Address Postcode? & What is the postal code of the address? & ST4 8AW & \ref{fig:app-charity-sample} \\
        \hline
        What is Address Street Line? & What is the value of the Address Street Line? & 28 Greenway & \ref{fig:app-charity-sample} \\
        \hline
        What is Charity Name? & What is the name of the charity? & Lucas' Legacy - Childhood Brain Tumour Research & \ref{fig:app-charity-sample} \\
        \hline
        What is Charity Number? & What is the charity number? & 1167650 & \ref{fig:app-charity-sample}\\
        \hline
        \addlinespace[5pt] 
        \hline
        What is Jurisdiction? & In which state is the company registered? & Delaware & \ref{fig:app-nda-sample}\\
        \hline
        What is Party? & What is the name of the company? & Cisco Systems, Inc., & \ref{fig:app-nda-sample}\\
        \hline
        \addlinespace[5pt] 
        \hline
        What is Contract ID? & What is the contract ID number? & 711207 & \ref{fig:app-ad-buy-sample} \\
        \hline
        What is the Product? & What is the name of the product being advertised? & Q42020 Broadcast & \ref{fig:app-ad-buy-sample}\\
        \hline
        What is Property? & What is the property name? & KXLF & \ref{fig:app-ad-buy-sample} \\
        \hline
        What is Agency? & Who is the advertising agency? & Left Hook Communications & \ref{fig:app-ad-buy-sample} \\
        \hline
        What is Advertiser? & Who is the advertiser? & Bennett/Democrat/\newline Secretary of State & \ref{fig:app-ad-buy-sample} \\
        \hline
        What is Gross Amount? & What is the value of the gross amount? & \$3,020.00 & \ref{fig:app-ad-buy-sample} \\
        \hline
        What is Sub Amount for M-F 530-7am News M-F 530-7am News? & What is the value for the 'Sub Amount' key for 'M-F 530-7am News'? & \$100.00 & \ref{fig:app-ad-buy-sample} \\
        \hline
        What is the Channel for M-F 530-7am News M-F 530-7am News? & What is the value of the 'Channel' for the '530-7am News' broadcasted from Monday to Friday? & All & \ref{fig:app-ad-buy-sample} \\
        \hline
        What is Program Start Date for M-F 530-7am News M-F 530-7am News? & What is the start date for the M-F 530-7am News program? & 10/06/20 & \ref{fig:app-ad-buy-sample} \\
        \hline
        What is the Program End Date for M-F 530-7am News M-F 530-7am News? & What is the end date for the program 'M-F 530-7am News'? & 10/12/20 & \ref{fig:app-ad-buy-sample} \\
        \hline
        \addlinespace[5pt]
        \hline
        What is Registrant Name? & What is the name of the registrant? & KOREA TRADE PROMOTION CENTER & \ref{fig:app-registration-sample} \\
        \hline
        What is Registration Number? & What is the registration number for the company? & 1619 & \ref{fig:app-registration-sample} \\
        \hline
        What is Signer Title? & What is the signer's title? & DEPUTY DIRECTOR & \ref{fig:app-registration-sample} \\
        \hline
        \addlinespace[5pt]
        \hline
        First bubble in the HPA Axis? & First bubble in the HPA Axis? & Hypothalamus & \ref{fig:app-dude-sample} \\
        \hline
        What does CORT stand for in this document? & What does CORT stand for in this document? & Cortisol & \ref{fig:app-dude-sample} \\
        \hline
        Where does cortisol go after it is sent from the adrenal cortex? & Where does cortisol go after it is sent from the adrenal cortex? & Hypothalamus & \ref{fig:app-dude-sample} \\
        \hline
        \addlinespace[5pt]
        \hline
        What is Buyer information? & What is the name of the buyer? & Buyer :Nichole Harrington 8282 Kristie Lights South Loriburgh, PR 35228 US Tel:+(227)782-8066 Email:blackjames@\newline example.net Site:http://ruiz-bailey.com/ & \ref{fig:app-fatura-sample} \\
        \hline
        What is Date of purchase? & When was the purchase date? & Invoice Date: 30-Oct-1998 & \ref{fig:app-fatura-sample} \\
        \hline
        What is Due date? & What is the due date? & Due Date : 24-May-2020 & \ref{fig:app-fatura-sample} \\
        \hline
        What is Purchase order number? & What is the purchase order number value? & PO Number :72 & \ref{fig:app-fatura-sample} \\
        \hline
        What is Seller Address? & What is the seller's address? & Address:05866 Velazquez Mount North Diane, NJ 20651 US & \ref{fig:app-fatura-sample} \\
        \hline
        What is Total amount before tax and discount? & What is the value of the total amount before tax and discount? & SUB\_TOTAL : 293.47 \$ & \ref{fig:app-fatura-sample} \\
        \hline
        What is Tax? & What is the tax amount? & TAX:VAT (5.69\%): 16.70 \$ & \ref{fig:app-fatura-sample} \\
        \hline
        What is Title? & What is the key for the title information? & TAX INVOICE & \ref{fig:app-fatura-sample} \\
        \hline
        What is Total amount to be paid? & What is the value of the total amount to be paid? & BALANCE\_DUE : 305.39 \$ & \ref{fig:app-fatura-sample} \\
        \hline
        \addlinespace[5pt]
        \hline
        What is MANUFACTURER:? & What is the value of the manufacturer? & R. J. REYNOLDS & \ref{fig:app-funsd-sample} \\
        \hline
        What is BRAND NAME:? & What is the value of the brand name? & CARDINAL CIGARETTES (11 PACKINGS) & \ref{fig:app-funsd-sample} \\
        \hline
        What is OTHER INFORMATION:? & What is the value of OTHER INFORMATION? & SEE ATTACHMENT & \ref{fig:app-funsd-sample} \\
        \hline
        \addlinespace[5pt]
        \hline
        to whom is this letter written to? & to whom is this letter written to? & Mr. Rionda & \ref{fig:app-mp-sample} \\
        \hline
        when is the letter dated ? & when is the letter dated ? & October 18, 1940, & \ref{fig:app-mp-sample} \\
        \hline
        \addlinespace[5pt]
        \hline
        what is the auth. no. mentioned in the given form ? & what is the auth. no. mentioned in the given form ? & 5754 & \ref{fig:app-sp-sample} \\
        \hline
        what is the value of percent per account as mentioned in the given form ? & what is the value of percent per account as mentioned in the given form ? & 50.06 & \ref{fig:app-sp-sample} \\
        \hline
        what is the emp. no. mentioned in the given form ? & what is the emp. no. mentioned in the given form ? & 483378 & \ref{fig:app-sp-sample} \\
        \hline
        what is the employee name mentioned in the given form ? & what is the employee name mentioned in the given form ? & IRENE KARL & \ref{fig:app-sp-sample} \\
        \hline
        what is the value of amount authorized per account ? & what is the value of amount authorized per account ? & 292.00 & \ref{fig:app-sp-sample} \\
        \hline
        \addlinespace[5pt]
        \hline
        Qual è Cognome? & Qual è Cognome? & ANNI & \ref{fig:app-xfund-sample} \\
        \hline
        Qual è Nome? & Qual è Nome? & GIACCOMO & \ref{fig:app-xfund-sample} \\
        \hline
        \hline
        Qual è Data Nascita? & Qual è Data Nascita? & 12/02/1988 & \ref{fig:app-xfund-sample} \\
        \hline
        Qual è Data? & Qual è Data? & 19/12/2020 & \ref{fig:app-xfund-sample} \\
        \hline
        Qual è Ora? & Qual è Ora? & 14:00 & \ref{fig:app-xfund-sample} \\
        \hline
        \caption{QA pairs of the examples, each pair referencing the specific example it corresponds to.}
        \label{tab:app_samples} \\
    \end{longtable}
\end{center}
\twocolumn

\begin{figure*}[ht]
    \centering
    \fbox{\includegraphics[height=0.95\textheight]{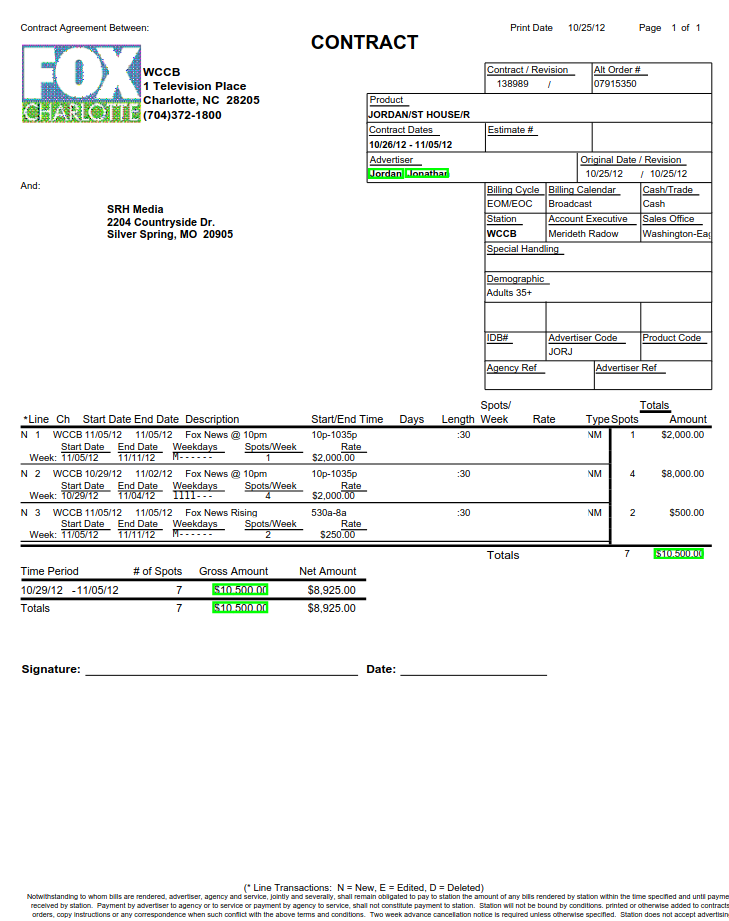}}
    \caption{\texttt{Deepform} sample}
    \label{fig:app-deepform-sample}
\end{figure*}

\begin{figure*}[ht]
    \centering
    \fbox{\includegraphics[height=0.95\textheight]{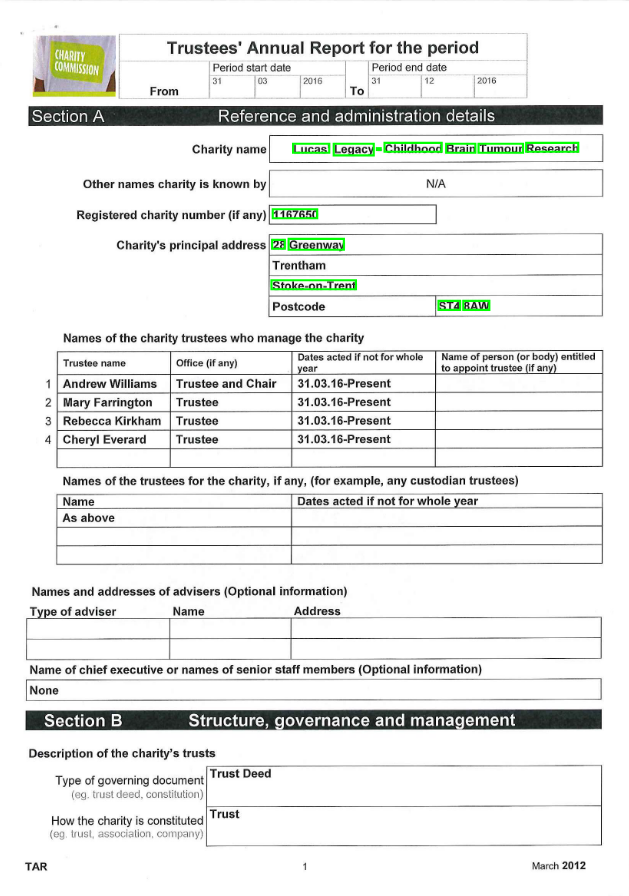}}
    \caption{\texttt{Kleister Charity} sample}
    \label{fig:app-charity-sample}
\end{figure*}

\begin{figure*}[ht]
    \centering
    \fbox{\includegraphics[height=0.95\textheight]{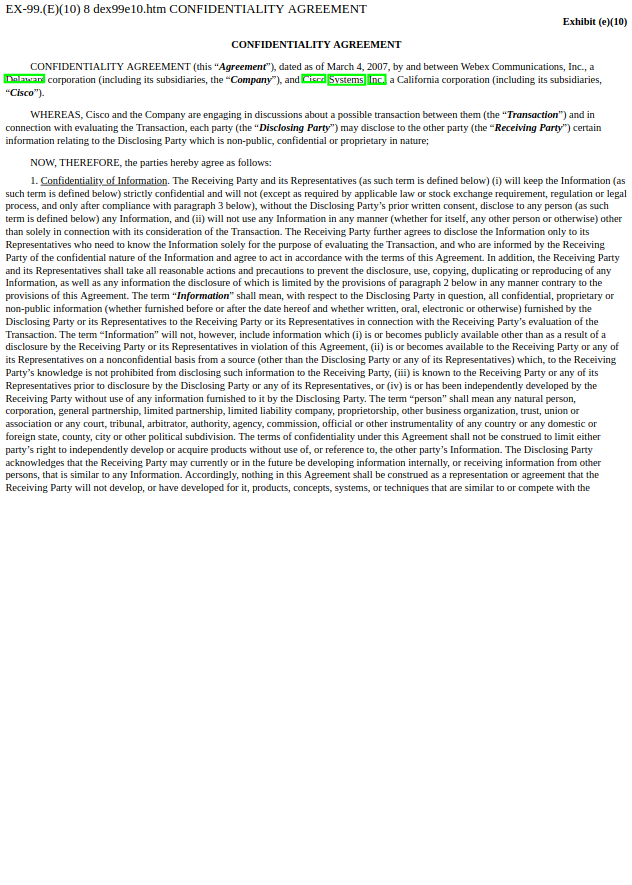}}
    \caption{\texttt{Kleister NDA} sample}
    \label{fig:app-nda-sample}
\end{figure*}

\begin{figure*}[ht]
    \centering
    \fbox{\includegraphics[height=0.95\textheight]{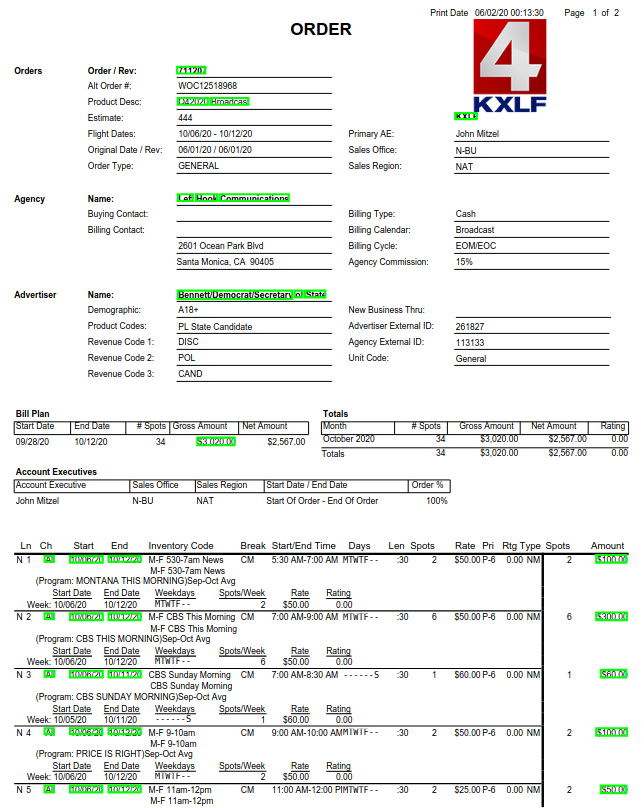}}
    \caption{\texttt{VRDU Ad Buy Form}. Not all the questions for this page are listed in Table \ref{tab:app_samples}, only until the details of the first broadcasting.}
    \label{fig:app-ad-buy-sample}
\end{figure*}

\begin{figure*}[ht]
    \centering
    \fbox{\includegraphics[height=0.95\textheight]{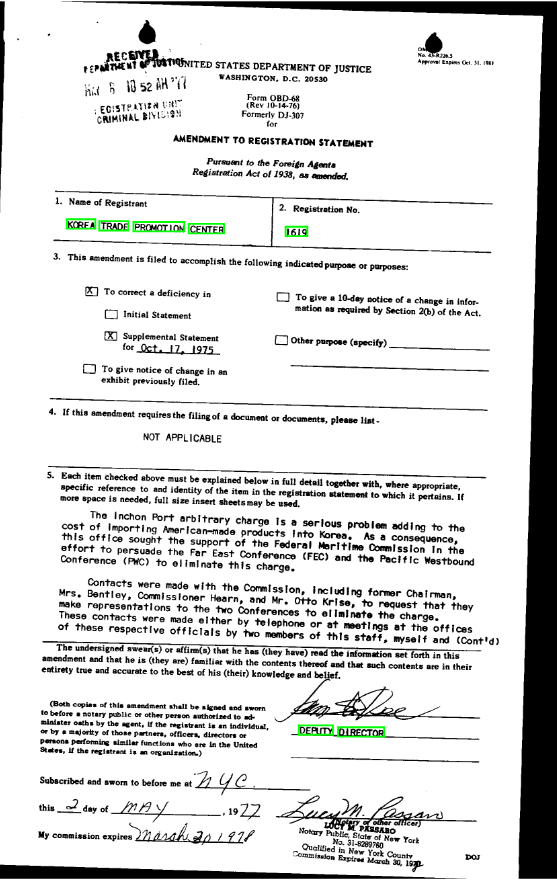}}
    \caption{\texttt{VRDU Registration Form} sample.}
    \label{fig:app-registration-sample}
\end{figure*}

\begin{figure*}[ht]
    \centering
    \fbox{\includegraphics[width=\textwidth]{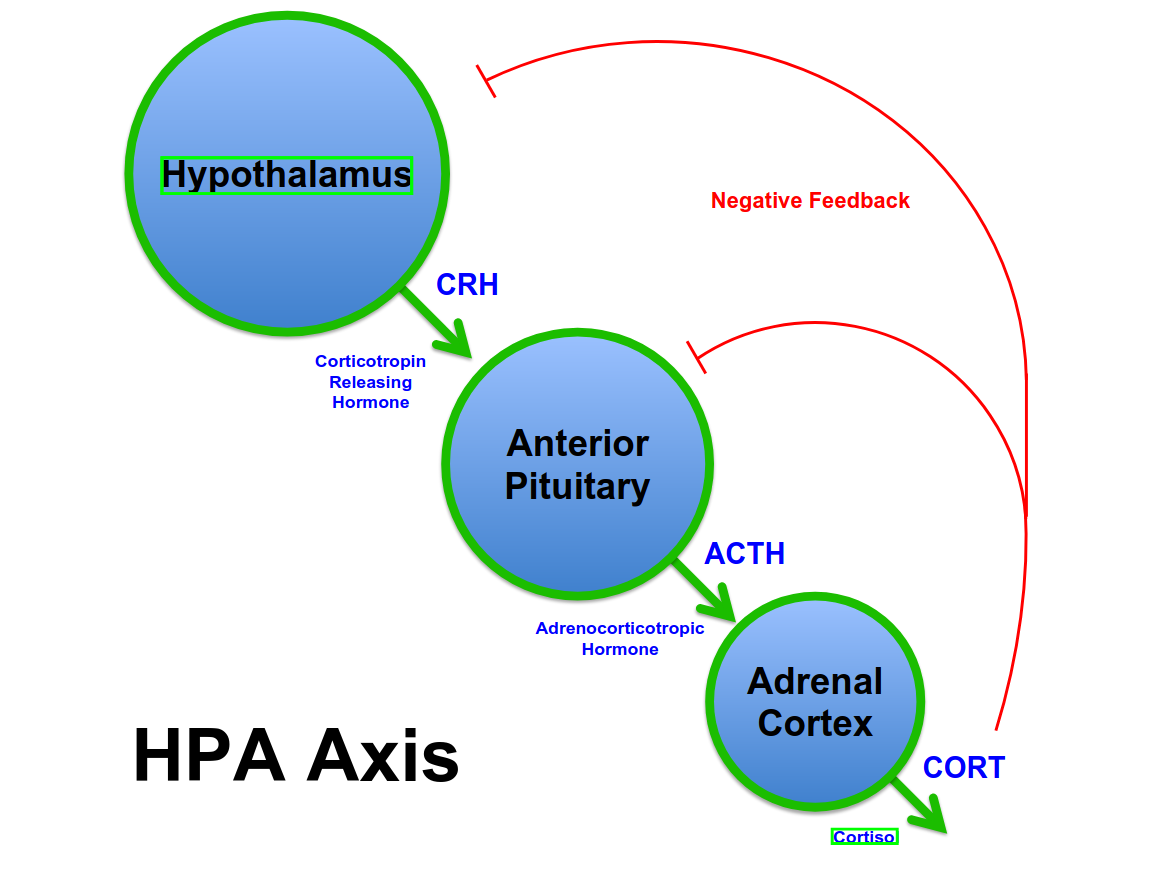}}
    \caption{\texttt{DUDE} sample.}
    \label{fig:app-dude-sample}
\end{figure*}

\begin{figure*}[ht]
    \centering
    \fbox{\includegraphics[height=0.95\textheight]{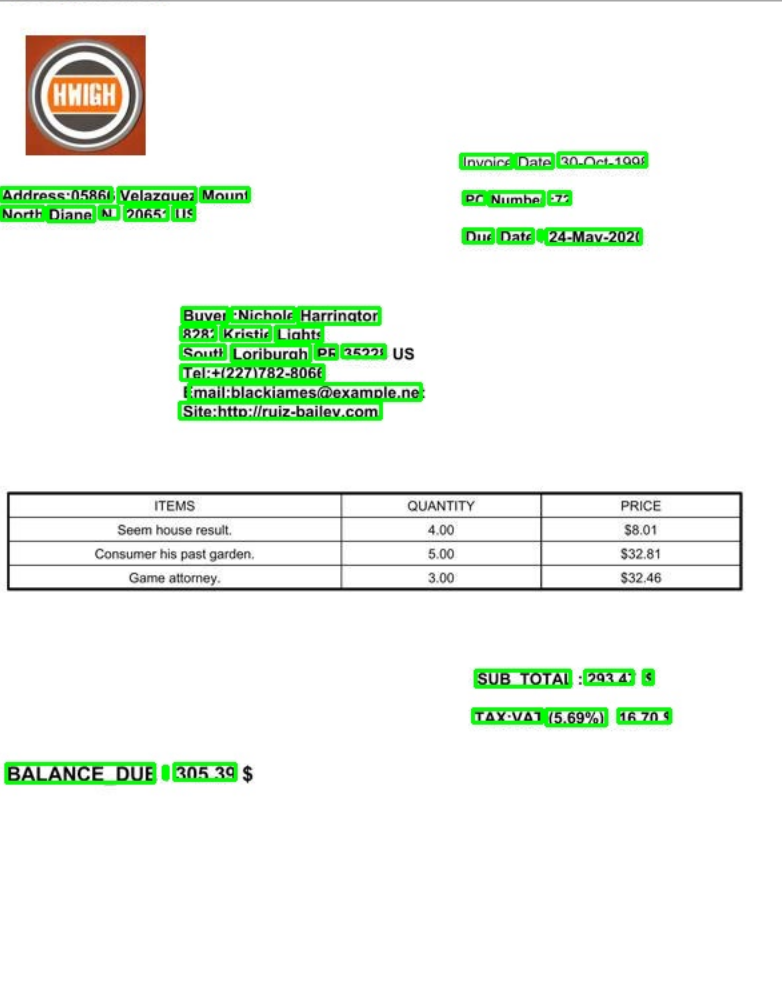}}
    \caption{\texttt{FATURA} sample.}
    \label{fig:app-fatura-sample}
\end{figure*}

\begin{figure*}[ht]
    \centering
    \fbox{\includegraphics[height=0.95\textheight]{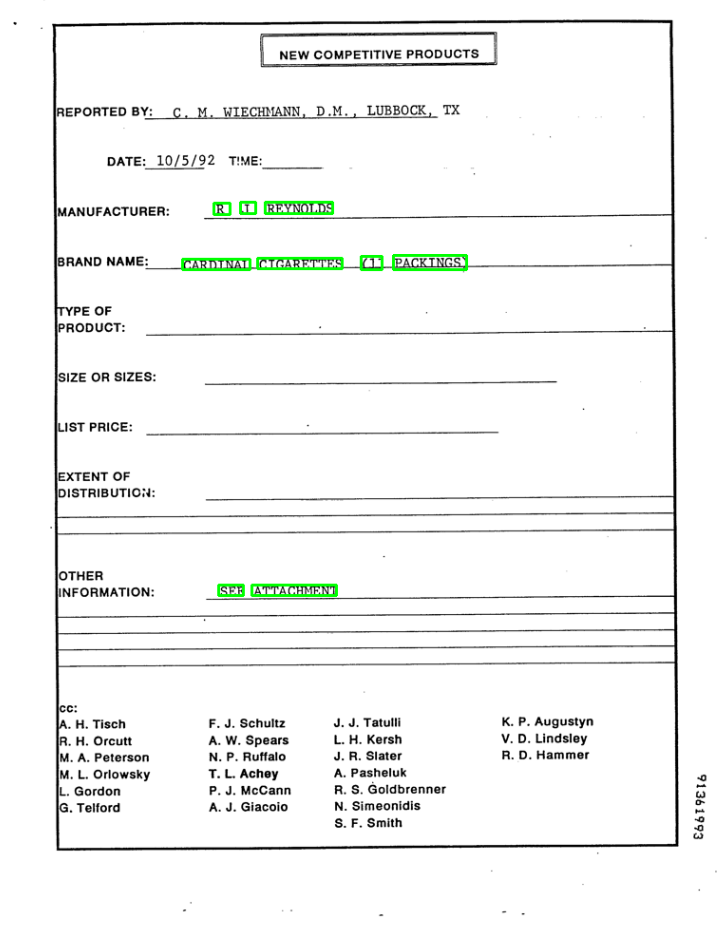}}
    \caption{\texttt{FUNSD} sample.}
    \label{fig:app-funsd-sample}
\end{figure*}

\begin{figure*}[ht]
    \centering
    \fbox{\includegraphics[height=0.95\textheight]{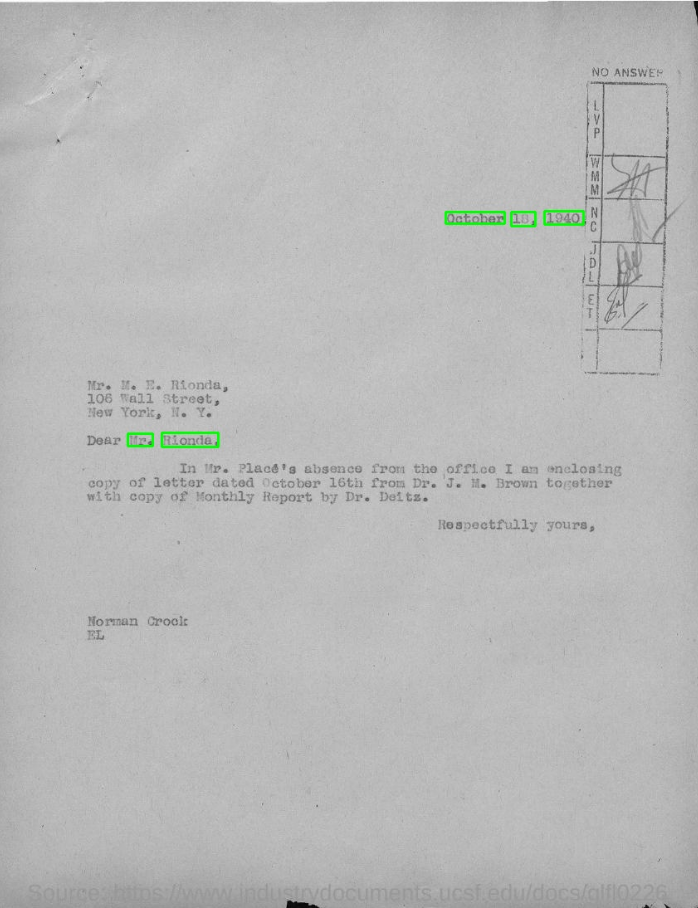}}
    \caption{\texttt{MP-DocVQA} sample.}
    \label{fig:app-mp-sample}
\end{figure*}

\begin{figure*}[ht]
    \centering
    \fbox{\includegraphics[width=\textwidth]{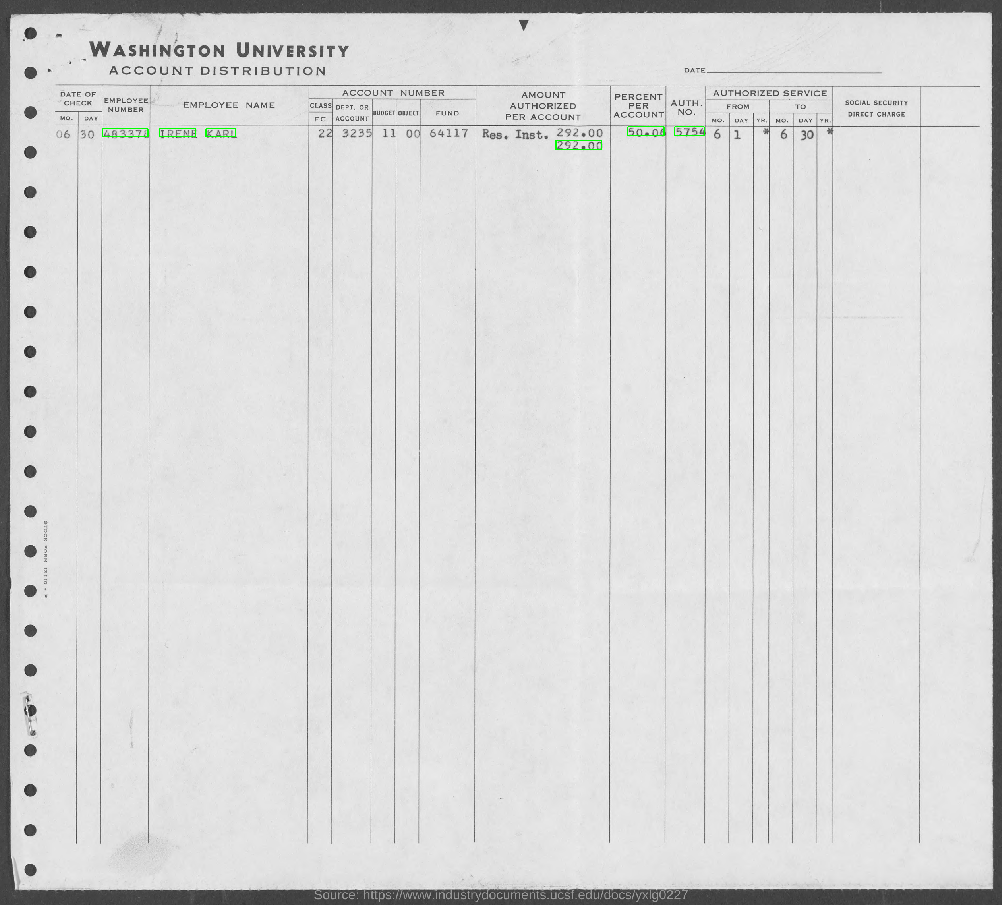}}
    \caption{\texttt{SP-DocVQA} sample.}
    \label{fig:app-sp-sample}
\end{figure*}

\begin{figure*}[ht]
    \centering
    \fbox{\includegraphics[height=0.95\textheight]{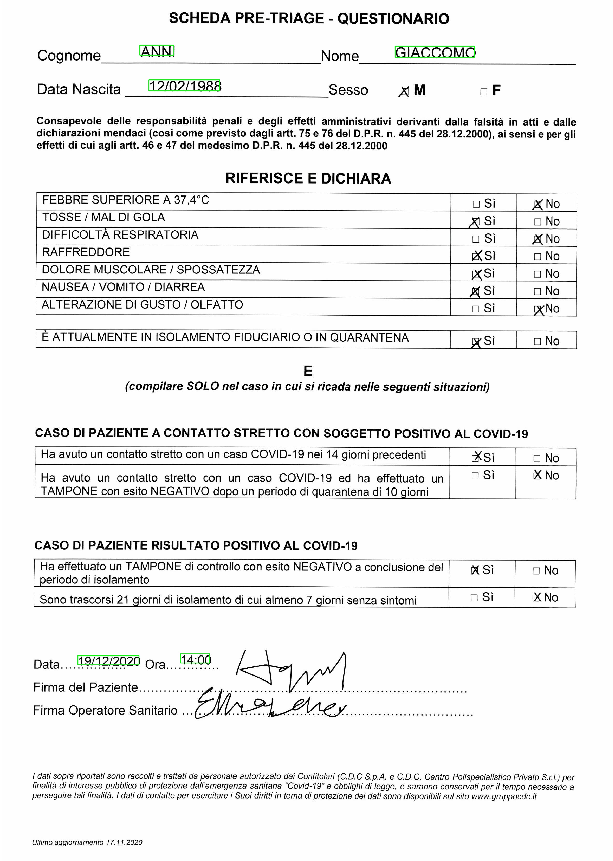}}
    \caption{\texttt{XFUND} sample.}
    \label{fig:app-xfund-sample}
\end{figure*}

%========================================

\end{document}